\documentclass[lettersize,journal]{IEEEtran}
\usepackage{hyperref}
\hypersetup{
	colorlinks=true,
	linkcolor=red,
	filecolor=magenta,      
	urlcolor=magenta,
	pdftitle={Overleaf Example},
	pdfpagemode=FullScreen,
} 
\usepackage{soul}
\usepackage[]{changes}
\usepackage{amsmath,amsfonts}
\usepackage{algorithmic}
\usepackage{algorithm}
\usepackage{array}
\usepackage[caption=false,font=normalsize,labelfont=sf,textfont=sf]{subfig}
\usepackage{parskip}
\usepackage{textcomp}
\usepackage{stfloats}
\usepackage{url}
\usepackage{verbatim}
\usepackage{graphicx}
\usepackage{cite}
\usepackage{graphicx}
\usepackage{booktabs}
\usepackage{multirow}
\usepackage{color}
\usepackage{booktabs}  
\usepackage{threeparttable}  
\usepackage{bigstrut}
\usepackage{svg}
\usepackage{makecell}
\svgsetup{
    inkscapepath=i/svg-inkscape/
}
\svgpath{{svg/}}
\usepackage{epstopdf}
\usepackage{indentfirst}
\usepackage{bibspacing}
\begin{document}
\title{Exploring Non-Local Spatial-Angular Correlations with a Hybrid Mamba-Transformer Framework for Light Field Super-Resolution}
\author{Haosong Liu~\IEEEmembership{}
\author{Haosong Liu, \emph{Graduate Student Member, IEEE}, Xiancheng Zhu, \emph{Graduate Student Member, IEEE},\\Huanqiang Zeng, \emph{Senior Member, IEEE}, Jianqing Zhu, \emph{Senior Member, IEEE}, Jiuwen Cao,  \emph{Senior Member, IEEE} and Junhui Hou, \emph{Senior Member, IEEE}
\vspace{-0.2cm}
\thanks{This work was supported in part by the Key Science and Technology Project of Fujian Province under Grant 2024HZ022007, in part by the Key Program of Natural Science Foundation of Fujian Province under Grant 2023J02022, in part by the Natural Science Foundation for Outstanding Young Scholars of Fujian Province under Grant 2022J06023, in part by the Natural Science Foundation of Fujian Province under Grant 2022J01294, in part by the Natural Science Foundation of Zhejiang Province under Grant LZ24F030010, and in part by the High-level Talent Team Project of Quanzhou City under Grant 2023CT001  (\emph{Corresponding author: Huanqiang Zeng}).}
\thanks{Haosong Liu is with the School of Information Science and Engineering, Huaqiao University, Xiamen 361021, China (e-mail: hsliu@stu.hqu.edu.cn).}
\thanks{Xiancheng Zhu is with the School of Mechanical Engineering and Automation, Huaqiao University, Xiamen 361021, China (e-mail: xianchengzhu@stu.hqu.edu.cn).}
\thanks{Huanqiang Zeng is with the School of Engineering, Huaqiao University, Quanzhou 362021, China, and also with the School of Optoelectronic and Communication Engineering, Xiamen University of Technology, Xiamen 361024, China (e-mail: zeng0043@hqu.edu.cn).}
\thanks{Jianqing Zhu is with the School of Engineering, Huaqiao University, Quanzhou 362021, China (e-mail: jqzhu@hqu.edu.cn).}
\thanks{Jiuwen Cao is with the Artificial Intelligence Institute, Hangzhou Dianzi University, Hangzhou 310018, China (e-mail: jwcao@hdu.edu.cn).}
\thanks{Junhui Hou is with the Department of Computer Science, City University of Hong Kong, Hong Kong (e-mail: jh.hou@cityu.edu.hk).}
\thanks{Haosong Liu and Xiancheng Zhu contributed equally to this work.}}
}
\markboth{}%
{Shell \MakeLowercase{\textit{et al.}}: A Sample Article Using IEEEtran.cls for IEEE Journals}
\maketitle
\begin{abstract}
Recently, Mamba-based methods, with its advantage in long-range information modeling and linear complexity, have shown great potential in optimizing both computational cost and performance of light field image super-resolution (LFSR). However, current multi-directional scanning strategies lead to inefficient and redundant feature extraction when applied to complex LF data. To overcome this challenge, we propose a Subspace Simple Scanning (Sub-SS) strategy, based on which we design the Subspace Simple Mamba Block (SSMB) to achieve more efficient and precise feature extraction. 
Furthermore, we propose a dual-stage modeling strategy to address the limitation of state space in preserving spatial-angular and disparity information, thereby enabling a more comprehensive exploration of non-local spatial-angular correlations. 
Specifically, in stage I, we introduce the Spatial-Angular Residual Subspace Mamba Block (SA-RSMB) for shallow spatial-angular feature extraction; in stage II, we use a dual-branch parallel structure combining the Epipolar Plane Mamba Block (EPMB) and Epipolar Plane Transformer Block (EPTB) for deep epipolar feature refinement. Building upon meticulously designed modules and strategies, we introduce a hybrid Mamba-Transformer framework, termed LFMT. LFMT integrates the strengths of Mamba and Transformer models for LFSR, enabling comprehensive information exploration across spatial, angular, and epipolar-plane domains. Experimental results demonstrate that LFMT significantly outperforms current state-of-the-art methods in LFSR, achieving substantial improvements in performance while maintaining low computational complexity on both real-word and synthetic LF datasets. The code is available at \href{https://github.com/hsliu01/LFMT}{https://github.com/hsliu01/LFMT}.
\end{abstract}
\vspace{-0.4cm}
\section{Introduction}
\begin{figure*}[th] 
	\centering
	\vspace{-0.2cm} 
	\includegraphics[width=1\linewidth]{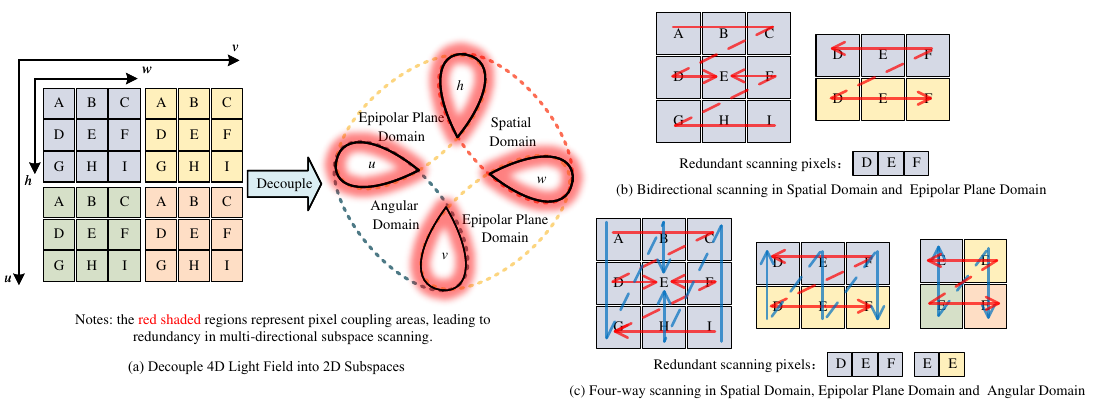}
	\caption{(a) The 4D LF can be decoupled into 2D subspaces, including the spatial, angular, and epipolar plane domains. For illustration, the LF images on the left is shown with \( u=v = 2 \) and \( h= w = 3 \), where different colors represent different viewpoints, and different letters indicate different pixels in the same viewpoint. (b) The coupling in bidirectional scanning\cite{39}: when bidirectional scanning is performed in spatial and epipolar plane domains, the correlations between blue pixels \( D \), \( E \), \( F \) are redundantly modeled. (c) The coupling in four-way scanning\cite{40}: when four-way scanning is performed in spatial, angular, and epipolar plane domains, the correlations between blue pixels \( D \), \( E \), \( F \), as well as between blue pixel \( E \), yellow pixel \( E \), are redundantly modeled.}
	\label{fig1}
	\vspace{-0.6cm}
\end{figure*}

\IEEEPARstart{L}{ight} Fields (LFs) capture both the intensities (spatial information) and directions (angular information) of light rays, providing richer information than traditional images \cite{1}. In recent years, the rise of portable LF cameras, such as Lytro and Raytrix, has accelerated the application of LF images, demonstrating great potential in areas such as post-capture refocusing \cite{2,3}, depth estimation \cite{4,5,6,7,7.1}, 3D reconstruction \cite{8}, and virtual reality \cite{9,10}. However, despite enhancing angular resolution through microlens arrays, these cameras inevitably compromise spatial resolution \cite{1}, thereby restricting their practical applicability. Therefore, LFSR algorithms have been widely studied to improve the spatial resolution of LF images.
\\\indent
In recent years, Convolutional Neural Networks (CNNs) have been widely applied to LFSR, aiming to learn complex spatial-angular information or correlations within LF images. A variety of feature extraction strategies have been proposed to alleviate the spatial-angular entanglement in 4D LF images.  Some methods \cite{33,11,12,13,14,T1,15,H1,17} extract features through several convolutions in single subspace to learn spatial or angular information, while others \cite{T2,T3,H2,H3,20,21,22,23} utilize combined convolutions across multiple subspaces to extract and fuse features, thereby learning spatial-angular correlations. However, these methods fail to fully leverage the complete information in all subspaces, and the local receptive field of CNNs restricts the interaction in individual subspace or across different subspaces, resulting in inadequate exploration of spatial-angular correlations.
\\\indent
Transformer models, known for their ability to capture long-range dependencies, have been increasingly applied to LFSR. Enhancing long-range interactions improves the modeling of non-local spatial-angular correlations. For example, networks such as LFT \cite{24}, DPT \cite{25} and LF-DET \cite{27} extract long-range information from spatial and angular domains, while EPIT \cite{28} alternately extract the horizontal and vertical structural information from epipolar plane domain. However, due to the quadratic complexity of self-attention computation, Transformer-based methods face challenges in constructing deeper networks to extract high-frequency information in all domains. This limitation hinders their ability to effectively and comprehensively explore non-local spatial-angular correlations in LF images. Therefore, effectively utilizing the complete information in all domains to model non-local spatial-angular correlations remains a significant challenge.
\\\indent
\IEEEpubidadjcol 
Recently, State Space Models (SSMs), particularly Mamba \cite{31}, have emerged as a powerful competitor to Transformer models, maintaining linear computational complexity while enabling long-range information modeling. Pure Mamba-based methods (e.g., MLFSR \cite{29} and LFMamba \cite{30}) adopt the multi-directional scanning strategy to model non-local spatial-angular correlations. However, they overlook the inherent complexity and redundancy of LF data, resulting in limited efficiency and insufficient capability to capture the fine-grained spatial-angular structures required for high-fidelity LFSR. In contrast, pure Transformer models excel at global context modeling and can better restore spatial-angular structures close to the original LF. Nevertheless, their reliance on unfolding 4D LF data into 1D sequences for modeling incurs prohibitive computational costs, thereby constraining overall performance. As illustrated in Fig. \ref{fig1}(b) and Fig. \ref{fig1}(c), the coupling of LF subspace scanning may lead to computational redundancy and degrade performance in modeling non-local spatial–angular correlations. To overcome this challenge, we design the Subspace Simple Mamba Block (SSMB) as a basic component for feature extraction in LF subspaces, which adopts the Subspace Simple Scanning (Sub-SS) strategy to mitigate the inefficiency and coupling (refer to Sec. \ref{Section3.c} for details). Additionally, to further enhance the utilization of feature space information, we introduce a symmetric branch composed of convolution and SiLU activation to compensate for the information loss caused by the Sub-SS sequential constraints.\\\indent
Furthermore, when the sequence length is excessively long,
the limited state space may fail to adequately preserve spatial-angular and disparity information, thereby hindering the
deeper exploration of spatial-angular correlations. To tackle this challenge, we propose a dual-stage modeling
strategy to capture non-local spatial-angular correlations. Specifically, in Stage I, we focus on the coarse extraction of
shallow spatial-angular features from spatial and angular domains. We first construct the Residual Subspace Mamba Block (RSMB) to reduce channel redundancy and achieve more efficient nonlocal feature aggregation. Subsequently, we design the Spatial-Angular Residual Subspace Mamba Block (SA-RSMB), where
the RSMB is ingeniously applied in the alternating processing
of the spatial and angular domains to capture non-local spatial
and angular feature. In Stage II, we thoroughly explore the deep interaction between LF disparity and structural information in epipolar plane domain for deep epipolar feature refinement. Building on the design strategy of SA-RSMB, we introduce the Epipolar Plane Mamba Block (EPMB) to
extract disparity and structural information from epipolar plane
domain, further refining the spatial-angular feature.
\\\indent
The epipolar plane domain deeply couples spatial and angular
information, as shown in Fig. \ref{fig1}(a). This information reflects the intrinsic structure of LF and contributes to promoting the refinement and enhancement of non-local spatial-angular correlations. Although EPMB demonstrates efficient long-range dependency modeling, its limited state space restricts deeper feature understanding, which may obstruct the capture of structural information in epipolar plane domain. To address this, we introduce the LFMT network that combines the strengths of Mamba and Transformer models for the first time. Specifically, we adopt a dual branch parallel structure, introducing a few Epipolar Plane Transformer Blocks (EPTB) with quadratic computational complexity in Stage II. This strategy compensates for the limitation of EPMB in enhancing spatial-angular correlations, enabling a more comprehensive exploration of deep disparity and structural information.\\\indent
We conduct extensive ablation experiments on real-world and synthetic LF datasets, validating the superior performance of LFMT in LFSR. Our contributions are summarized as follows:
\begin{itemize}
\item We propose a Sub-SS strategy and construct the SSMB, which addresses the issue of inefficient feature extraction caused by the coupling of LF subspace scanning in LFSR.
\item We propose a dual-stage strategy for modeling non-local spatial-angular correlations that effectively preserves spatial-angular and disparity information.
\item We propose the LFMT network, which combines the advantages of Mamba and Transformer models for LFSR. LFMT accomplishes non-local spatial-angular correlations modeling from coarse to fine through complete information interaction in all 2D subspaces.
\item Experimental results confirm that our method achieves new state-of-the-art performance in LFSR both quantitatively and qualitatively, while maintaining low computational complexity.
\end{itemize}

\section{Relate Work}\label{Section2}
\subsection{Light Field 2D Subspace Representation}
\begin{figure}[th]
	\centering
	\vspace{-0.2cm}
	\includegraphics[width=0.5\textwidth, height=0.5\textheight,keepaspectratio]{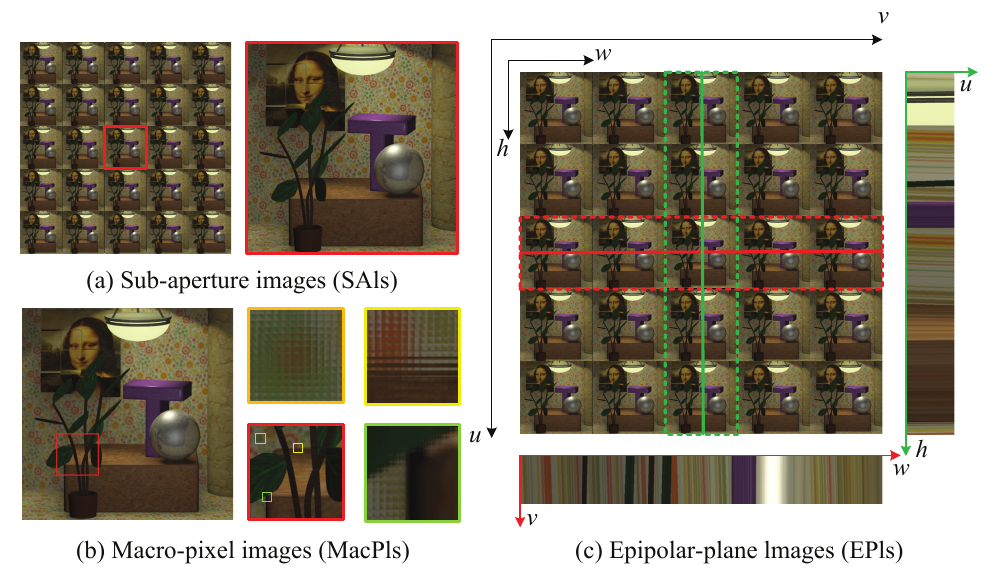}
	\caption{The projection of 4D LF data into 2D subspaces: (a) SAI pattern projected into the spatial domain, (b) MacPI pattern projected into the angular domain, and (c) EPI pattern projected into the epipolar plane domain.}
	\label{fig2}
\end{figure}

LFs are commonly represented in 4D using the two-plane parameterization model $L(u, v, h, w)$ proposed by Levoy et al. \cite{c1}, where $(u, v)$ represents the viewpoint plane, and $(h, w)$  represents the image plane. However, the spatial and angular information in the complex and voluminous 4D LF data are highly coupled, making the modeling of spatial-angular correlations challenging. A common approach is to project 4D LF data into 2D subspaces, as shown in Fig. \ref{fig2}. The primary types of 2D subspace representations include sub-aperture images (SAIs), macro-pixel images (MacPIs), and epipolar-plane images (EPIs), respectively corresponding to the spatial, angular, and epipolar-plane domains. The epipolar-plane domain bridges the spatial and angular domains, encoding spatial-angular information. Notably, EPIs reflect disparity values through slanted line patterns, enabling the exploration of deeper disparity and structural information.
\subsection{Light Field Image Super-Resolution}
The task of LFSR typically aims to enhance the spatial resolution of each sub-aperture image. As LF images become increasingly adopted in various applications, this technology has received considerable attention in recent years. By utilizing advancements in modern neural networks, LFSR has made significant breakthroughs. Currently, mainstream deep learning-based methods for LFSR can be broadly categorized into two types: CNN-based methods and Transformer-based methods.\\\indent
To address the challenges in spatial-angular feature modeling for LFSR, various CNN-based methods have been proposed with progressive improvements. Yoon \textit{et al.} \cite{33} first introduced LFCNN, pioneering the use of CNNs to explore correlations between adjacent SAIs and achieving joint spatial-angular super-resolution. Building upon this, Zhang \textit{et al.} \cite{11} proposed resLF, which groups SAIs with consistent sub-pixel disparities into stacks and employs residual convolutions to extract epipolar features more effectively. To further enhance spatial-angular interaction, Yeung \textit{et al.} \cite{12} introduced 4D convolutions and spatial-angular separated convolutions to model complex structural dependencies in both domains. Jin \textit{et al.} \cite{13} presented the ``All-to-One'' model, integrating complementary information across multiple viewpoints while preserving the disparity structure of LF images. Extending the separation and fusion paradigm, Wang \textit{et al.} \cite{18} developed LF-InterNet, which extracts spatial and angular features independently and fuses them via feature interaction modules. This design was further improved in DistgSSR \cite{19}, which introduces a decoupling mechanism based on convolutions with varying dilation rates to more flexibly capture spatial-angular correlations. The works of Hou \textit{et al.} \cite{H4,H5,H6} make significant contributions to LFSR by introducing complementary strategies such as view synthesis, coded-aperture camera reconstruction, and regularized feature learning, which enhance the recovery of both spatial and angular information in a more comprehensive manner. To robustly handle diverse real-world degradations, Xiao \textit{et al.} \cite{62} proposed LF-DEST, a blind LFSR method that integrates explicit degradation estimation with a modulated selective fusion module. More recently, Van \textit{et al.} \cite{23} proposed HLFSR, which decomposes the 4D LF into multiple 2D subspaces to enable efficient feature extraction and attention-based fusion. Despite their evolution from early pairwise SAI modeling to advanced exploitation of epipolar geometry and subspace decomposition, CNN-based methods remain constrained by limited receptive fields, hindering effective modeling of non-local spatial-angular correlations.\\\indent
Recent studies have introduced Transformer-based methods into LFSR to solve the problem of global information interaction. The main idea of these methods is to transform pixels into sequences and approach LFSR as a sequence-to-sequence reconstruction task. By the self-attention mechanism, these methods can learn non-local spatial-angular correlations and model long-range dependency across the spatial and angular domains. Liang \textit{et al.} \cite{24} first proposed LFT and utilised Transformer-based architecture to alternately extract long-range feature in spatial and angular domains, achieving non-local spatial-angular information interaction. Wang \textit{et al.} \cite{25} introduced the detail-preserving transformer, which flattens vertical or horizontal SAIs and employs LF gradient maps to guide sequence learning, thereby effectively enhancing detail restoration. Liang \textit{et al.} \cite{28} further developed EPIT, a robust network for the large disparity LF scenes. They input vertical and horizontal EPI pixels as sequences and leveraged basic transformer to learn non-local spatial-angular correlations. Cong \textit{et al.} \cite{27} proposed LF-DET, which applies subsampling convolution layers in spatial domain for sequence mapping and adopts a multi-scale angular modeling strategy in angular domain, achieving more efficient information processing. Xiao \textit{et al.} \cite{61} proposed OHT, a hybrid network that combines convolution and Transformer modules with occlusion-aware spatial-angular modeling to efficiently capture local and global correlations. However, Transformer-based methods face the quadratic computational complexity. This leads to challenges in efficiently applying transformer blocks in all 2D subspaces and in constructing deep networks to extract high-frequency information. Beyond network architecture design, Xiao \textit{et al.} \cite{63}  introduced CutMIB, a novel data augmentation strategy for LFSR that leverages implicit geometric cues via multi-view patch blending to enhance reconstruction accuracy and angular consistency.
\subsection{State Space Model}
SSMs originate from classical control theory \cite{35} and have recently emerged as a promising architecture in deep learning for long-range dependency modeling. SSMs capture long-range dependency through linear recurrence processes with a computational complexity of $O(N)$, offering a significant advantage over the $O(N^{2})$ complexity of Transformer-based methods. S4 \cite{36} introduced a novel parameterization method for structured state spaces, significantly improving the efficiency of long sequence modeling. Building on this, S5 \cite{37} further enhanced long sequence processing capabilities through the multi-input multi-output mechanism and efficient parallel scanning. The H3 model \cite{38} narrowed the performance gap between SSMs and Transformer in natural language tasks, demonstrating the huge potential of SSMs in various applications. In particular, Mamba \cite{31} introduced a selective mechanism that adaptively learns structured parameters from input data, greatly enhancing the model's performance.\\\indent
Mamba-based methods have been rapidly applied to vision tasks due to their strengths in linear scalability, computational efficiency, and flexibility. The initial studies, including Vim \cite{39} and VMamba \cite{40}, employed multi-directional scanning to overcome the causality limitations of Mamba. These works have inspired the application of Mamba in a range of vision tasks, including image segmentation \cite{41}, image restoration \textit{et al.}, video understanding \cite{46,47} and 3D point cloud processing \cite{48,49}. Mamba’s unique strengths in long-range dependency modeling and computational efficiency establish it as a promising method for LFSR. Specifically, Gao \textit{et al.} \cite{29} introduced bidirectional scanning with Mamba for LFSR. Xia \textit{et al.} \cite{30} proposed four-directional scanning along the channel dimension to reduce the excessive number of parameters of Mamba for LFSR task. Additionally, Jin \textit{et al.} \cite{64} proposed the LFTransMamba that integrates Mamba and Transformer, combined with masked LF image modeling and an enhanced position-sensitive window mechanism, to balance spatial–angular consistency modeling and computational complexity. However, pure Mamba-based methods suffer from the coupling of 2D subspace scanning in LF, where pixels are redundantly modeled across subspaces, thus constraining performance. In this work, we further exploit the potential of Mamba for LFSR by introducing a hybrid Mamba-Transformer architecture, which enables comprehensive information interaction across all 2D subspaces and facilitates non-local spatial-angular correlation modeling in a coarse-to-fine manner.

\begin{figure*}[t]
\centering
\vspace{-0.1cm}
\includegraphics[width=1\linewidth]{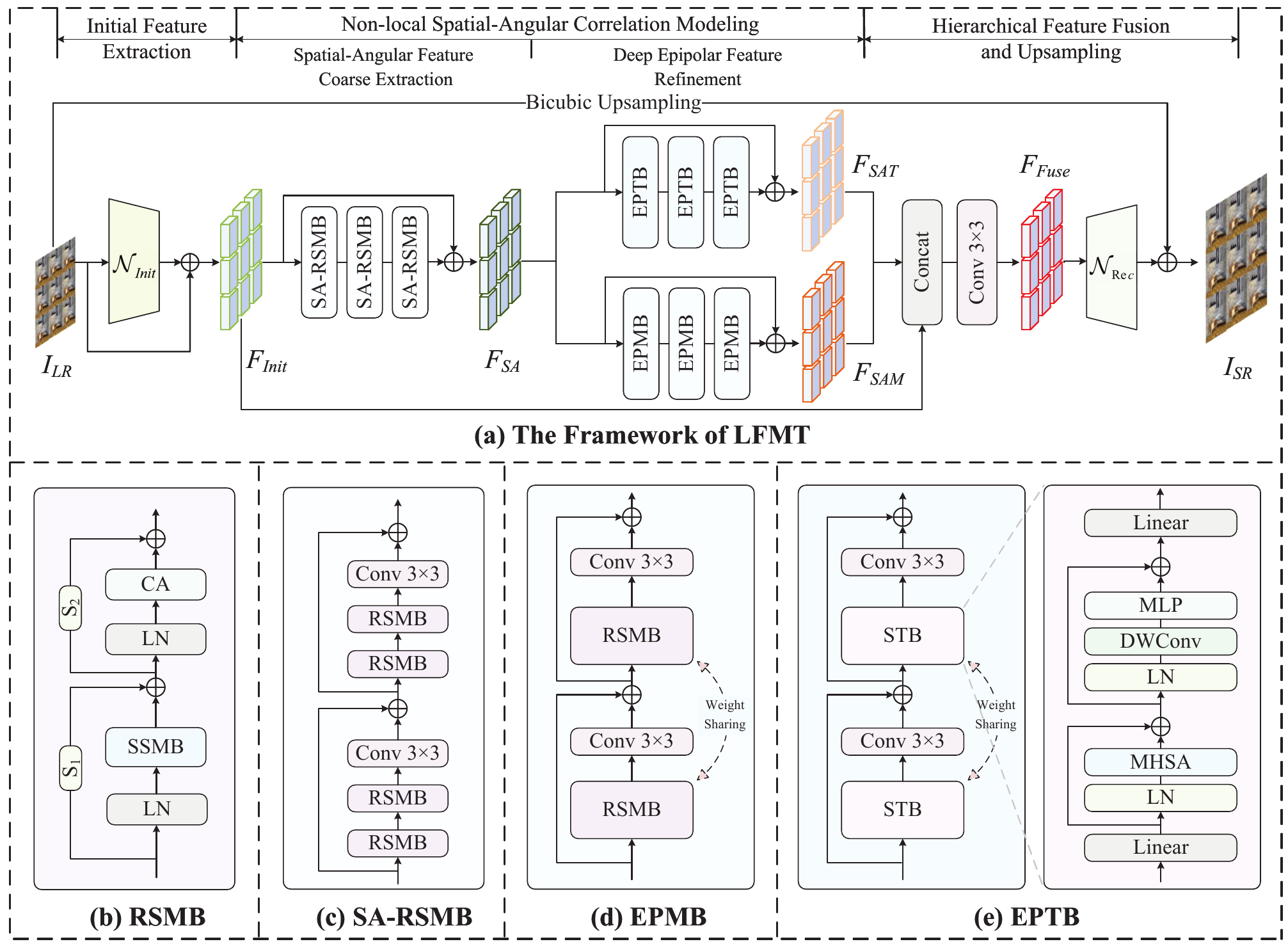}
\vspace{-0.5cm}
\caption{(a) The overall architecture of the proposed LFMT network. (b) RSMB is used for effective long-range dependency modeling. (c) SA-RSMB is used for shallow spatial-angular feature coarse extraction in spatial and angular domains. (d) EPMB and (e) EPTB are used for deep refinement and enhancement of spatial-angular feature in epipolar plane domain.}
\label{fig3}
\vspace{-0.4cm}
\end{figure*}

\section{Method} \label{Section3}
\subsection{Mamba Preliminaries}
SSMs are the mathematical framework widely used in time series analysis and control systems. They describe the evolution of the system's hidden states and internal dynamic relationships through state equations. The core idea is to establish relationships between the input sequence \(x(t) \in \mathbb{R}^L\) and the output sequence \(y(t) \in \mathbb{R}^L\) through intermediate hidden states \(h(t) \in \mathbb{R}^L\), which capture the underlying structure and dynamic properties of the system. This process is typically represented by linear ordinary differential equations :
\begin{equation}
	\begin{aligned}
		 {h'}(t) &= \mathbf{A} h(t) + \mathbf{B} x(t), \\
		 y(t) &= \mathbf{C} h(t),
	\end{aligned}
\label{eq1}
\end{equation}
where \(L\) is the length of the input sequence, \(\mathbf{A} \in \mathbb{R}^{N \times N}\) is the state matrix, \(\mathbf{B} \in \mathbb{R}^{N \times 1}\), and \(\mathbf{C} \in \mathbb{R}^{N \times 1}\) are projection parameters.\\\indent 
In recent years, for adapt to the training of deep neural networks, the structured state space sequence model (S4) \cite{36} has introduced a timescale parameter \(\Delta\) to discretize the process in Eq. \eqref{eq1}. They use the zero-order hold rule to convert continuous parameters \(\mathbf{A}\) and \(\mathbf{B}\) into discrete parameters \(\overline{\mathbf{A}}\) and \(\overline{\mathbf{B}}\):
\begin{equation}
	\begin{aligned}
		& \overline{\mathbf{A}} = \exp(\Delta \mathbf{A}), \\
		& \overline{\mathbf{B}} = (\Delta \mathbf{A})^{-1} (\exp(\Delta \mathbf{A}) - \mathbf{I}) \cdot (\Delta \mathbf{B}).
	\end{aligned}
\end{equation}
\\\indent Thus, Eq. \eqref{eq1} is derived into its discrete form:
\begin{equation}
	\begin{aligned}
		& h_k = \overline{\mathbf{A}} h_{k-1} + \overline{\mathbf{B}} x_k, \\
		& y_k = \mathbf{C} h_k.
	\end{aligned}
 \label{eq3}
\end{equation} \\\indent 
Further, the recursive form of Eq. \eqref{eq3} can be converted into a fast parallel training structure with the structured convolution kernel \(\overline{\mathbf{K}} \in \mathbb{R}^L\) :
\begin{equation}
	\begin{aligned}
		& \overline{\mathbf{K}} = (\mathbf{C} \overline{\mathbf{B}}, \mathbf{C} \overline{\mathbf{A}} \overline{\mathbf{B}}, \dots, \mathbf{C} \overline{\mathbf{A}}^{L-1} \overline{\mathbf{B}}), \\
		& y = x * \overline{\mathbf{K}},
	\end{aligned}
\end{equation}
where \(*\) denotes the convolution operation. \\\indent 
It bears mentioning that Mamba \cite{31} introduces a context-aware information selection mechanism on the parameters \(\overline{\mathbf{B}}\) and \(\mathbf{C}\) and proposes a parallel scanning algorithm to accelerate training. This approach achieves weight allocation related to the input and maintains linear computational complexity at the same time, significantly enhancing the efficiency and performance of SSMs.
\subsection{Overview of LFMT}
For the LFSR task, given a low-resolution LF image \( {{I}_{LR}} \in \mathbb{R}^{U \times V \times H \times W} \), the goal is to reconstruct a high-resolution LF image \( {{I}_{HR}} \in \mathbb{R}^{U \times V \times \alpha H \times \alpha W} \), where \( U \times V \) is the angular resolution, \( H \times W \) is the spatial resolution and \( \alpha \) is the scale factor. As shown in Fig. \ref{fig3}(a), the LFMT consists of three parts: initial feature extraction, non-local spatial-angular correlations modeling, as well as hierarchical feature fusion and upsampling. Similar to previous works \cite{24,28}, we first use an encoder \( {\mathcal{N}}_{Init} \), composed of cascaded convolutionals, to extract initial features \( {{F}_{Init}} \in \mathbb{R}^{U \times V \times H \times W \times C} \) from the input \( {{I}_{LR}} \), where \( C = 64 \) is the channel dimension. Then, \( {{F}_{Init}} \) undergoes a dual-stage spatial-angular correlations modeling from coarse-grained to fine-grained, to accomplish non-local information interaction in all subspaces. Specifically, in Stage I, \( {{F}_{Init}} \) will be processed by three cascaded SA-RSMBs to extract spatial-angular features \( {{F}_{SA}} \in \mathbb{R}^{U \times V \times H \times W \times C} \) coarsely. In Stage II, \( {{F}_{SA}} \) will be processed by a dual-branch parallel structure which contains EPMBs and EPTBs, to respectively generate refined and enhanced spatial-angular features \( {{F}_{SAM}} \in \mathbb{R}^{U \times V \times H \times W \times C} \) and \( {{F}_{SAT}} \in \mathbb{R}^{U \times V \times H \times W \times C} \). Subsequently, \( {{F}_{Init}} \), \( {{F}_{SAM}} \), and \( {{F}_{SAT}} \) are fused to generate the multi-level features \( {{F}_{Fuse}} \in \mathbb{R}^{U \times V \times H \times W \times 3C} \) :
\begin{equation}
	{{F}_{Fuse}} = Conv(Concat({{F}_{Init}}, {{F}_{SAM}}, {{F}_{SAT}})),
\end{equation}
where \(Conv(\cdot)\) denotes the convolution operation, and \(Concat(\cdot)\) indicates concatenation along the channel dimension.
Finally, \( {{F}_{Fuse}} \) undergoes an upsampling encoder \( {\mathcal{N}}_{Rec} \), which consists of pixel shuffling layers and convolutional layers, to generate the final high-resolution image \( {{I}_{HR}} \).
\subsection{Subspace Simple Scanning and Subspace Simple Mamba Block}
\label{Section3.c}
\begin{figure}[]
	\centering
	\includegraphics[width=0.48\textwidth, height=0.5\textheight,keepaspectratio]{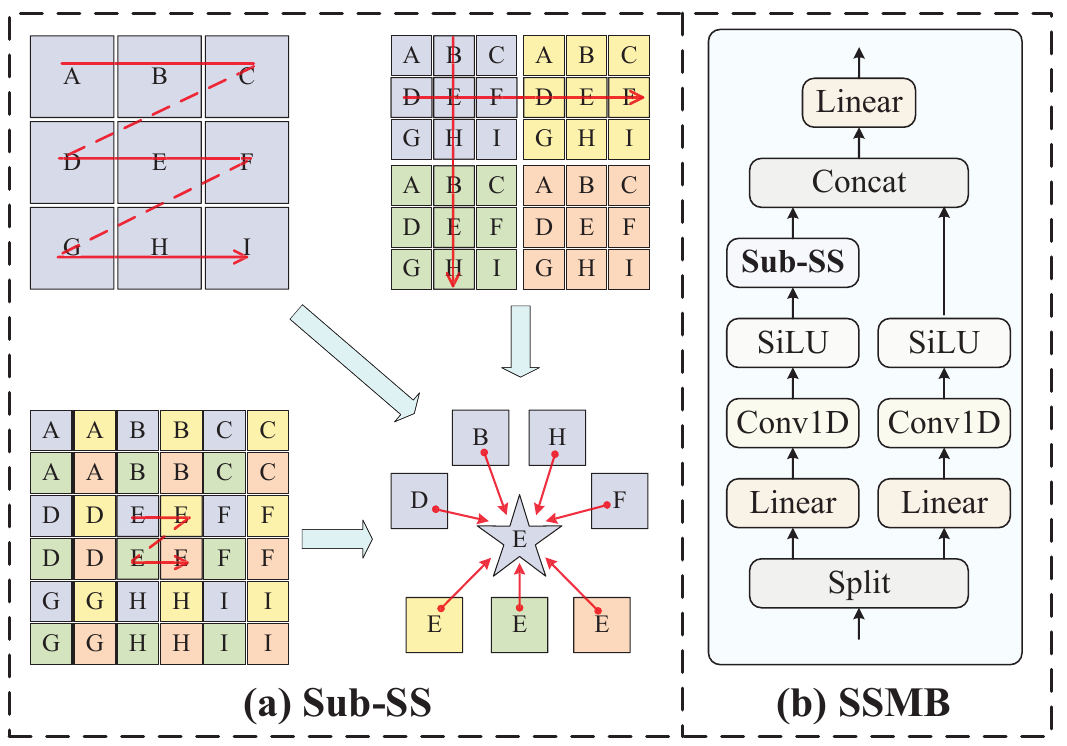}
	\vspace{0.1cm}
	\caption{(a) Sub-SS effectively models correlations between LF pixels and their nearest pixels in all subspaces through a simple unidirectional scanning. (b) The architecture of SSMB, which replaces causal convolution layers with regular convolution layers and introduces a symmetric path without SSM.}
	\label{fig4}
\end{figure}

Inspired by \cite{24,28}, we propose slicing 4D LF features into 2D subspaces, including the SAI pattern \(F_s \in \mathbb{R}^{H \times W \times C} \ (s = 1, \dots, UV)\), the MacPI pattern \(F_a \in \mathbb{R}^{U \times V \times C} \ (a = 1, \dots, HW)\), as well as the horizontal EPI pattern \(F_{hor} \in \mathbb{R}^{V \times W \times C} \ (hor = 1, \dots, UH)\) and vertical EPI pattern \(F_{ver} \in \mathbb{R}^{U \times H \times C} \ (ver = 1, \dots, VW)\). These features are then flattened through scanning. Furthermore, considering the coupling of LF subspace scanning, we design the Sub-SS strategy. Sub-SS has two core characteristics:  (1) it replaces the traditional multi-directional scanning used in visual state space models with a unidirectional scanning approach to model feature correlations, and (2) it scans across all subspaces to thoroughly capture the spatial-angular and geometric features of LF. As shown in Fig. \ref{fig4}(a), the Sub-SS strategy selects a specific pixel, referred to as the observation center (e.g., the blue pixel E). From this center, Sub-SS efficiently aggregates information from neighboring pixels in spatial domain and related pixels in angular domain, while also incorporating pixel disparity and geometric information from the epipolar plane. This process  utiliz surrounding pixels (e.g., the blue pixel B, D, F, H and pixels of different colors E) to gather relevant spatial-angular information essential for accurately modeling feature distributions. The unidirectional scanning simplifies data aggregation, reducing computational complexity while enhancing the relevance of the information collected. Additionally, Sub-SS captures angular relationships by analyzing aligned pixels in angular domain and integrates geometric structure from the epipolar plane to improve the model's understanding of spatial relationships. In summary, Sub-SS combines unidirectional scanning with comprehensive subspace information to effectively model complex correlations in LF images, streamlining computation and improving performance in capturing spatial-angular and geometric features.\\\indent 
Building upon Sub-SS, we design the SSMB to efficiently extract features. Specifically, as shown in Fig. \ref{fig4}(b), different from the normal mamba structure, we replace causal convolutions with regular convolutions to eliminate the unidirectional limitation. Additionally, a symmetric branch, composed of an extra convolutional layer and SiLU activation, is introduced to compensate for potential information loss caused by the sequential modeling in Sub-SS. Subsequently, the outputs of two branches are concatenated and projected through a linear layer. This process effectively fuses sequential and spatial information, leveraging the complementary advantages of both features. Notably, to control the parameter size, the output of each branch is mapped to an embedding space with \(C/2\) dimension. For the given input feature \(F_{in}\), the output feature \(F_{out}\) of SSMB can be described as the following computation process:
\begin{equation}
	\begin{aligned}
		F_1 &= SS\text{-}sub(\sigma (Conv(Linear(C, \frac{C}{2})(F_{in})))),  \\ 
		F_2 &= \sigma (Conv(Linear(C, \frac{C}{2})(F_{in}))),  \\ 
		F_{out} &= Linear(C, C)(Concat(F_1, F_2)),
	\end{aligned}
\end{equation}
where \(F_1\) and \(F_2\) denote the intermediate features, \(\sigma\) denotes the sigmoid-weighted linear unit activation function, and \(Linear(\cdot)\) denotes the linear mapping layer.

\subsection{Dual-stage Non-local Spatial-Angular Correlations Modeling Strategy}
The dual-stage modeling strategy can progressively model non-local spatial-angular correlations from coarse to fine. As shown in Fig. \ref{fig3}(b), similar to the approach in \cite{44}, we introduced a channel attention layer after SSMB to enhance interactions among channel features. Furthermore, a learnable residual mechanism was employed to construct RSMB. Specifically, for given features \(F\in {{\mathbb{R}}^{B\times H\times W\times C}}\), the RSMB captures long-range dependency in pixels and outputs an optimized feature representation \(\widehat{F}\in {{\mathbb{R}}^{B\times H\times W\times C}}\), formulated as follows: 
\begin{equation}
	\begin{aligned}
		\overline{F} &= SSMB(LN(F)) + s_1 \cdot F, \\ 
		\widehat{F} &= CA(LN(\overline{F})) + s_2 \cdot \overline{F}, 
	\end{aligned}
\end{equation}
where \(LN(\cdot)\) denotes layer normalization, \(CA(\cdot)\) represents the channel attention layer, and \(s_1\), \(s_2\) are learnable residual connections.
\subsubsection{Stage I -- SA-RSMB for shallow spatial-angular feature coarse extraction in spatial and angular domains}
Stage I, denoted as \( {{H}_{StageI}}(\cdot) \), is primarily designed to integrate contextual information in spatial domain and complementary angular information in angular domain, mapping \( F_{Init} \) to \( F_{SA} \). To achieve this, three cascaded SA-RSMBs and a local skip connection are used:
\begin{equation}
\small
	\begin{aligned}
		F_{SA} &= {{H}_{StageI}}(F_{Init}) + F_{Init},\\ 	
		{{H}_{StageI}}(\cdot) &= SA\text{-}RSMB^3(SA\text{-}RSMB^2(SA\text{-}RSMB^1(\cdot)))
	\end{aligned}
\end{equation}
where \( SA\text{-}RSMB^i(\cdot), i \in [1,2,3] \) denotes \(i\)-th SA-RSMB.
\\\indent
The specific structure of SA-RSMB is shown in Fig. \ref{fig3}(c). We adopt a simple yet effective spatial-angular separable modeling method. First, the input feature \( F \in \mathbb{R}^{U \times V \times H \times W \times C} \) is reshaped into \( F_{SAI} \in \mathbb{R}^{UV \times H \times W \times C} \). It then passes through two RSMBs and a convolution for spatial domain feature interaction. This process integrates contextual information in spatial domain, generating the feature \( F_S \in \mathbb{R}^{U \times V \times H \times W \times C} \) that contains fused spatial domain information:
\begin{equation}
	F_S = Conv(RSMB^2(RSMB^1(F_{SAI}))) + F_{SAI}.
\end{equation}
\indent
Next, \( F_S \) is reshaped into \( F_{MacPI} \in \mathbb{R}^{HW \times U \times V \times C} \), and undergoes feature interaction in the angular domain via similar RSMBs and convolution. This process further utilizes complementary information in angular domain, generating the feature \( F_{SA} \in \mathbb{R}^{U \times V \times H \times W \times C} \):
\begin{equation}
  \small
  \begin{aligned}
	F_{SA} = Conv(RSMB^2(RSMB^1(F_{MacPI}))) + F_{MacPI},
  \end{aligned} 
\end{equation}
where \( RSMB^j(\cdot), j \in [1,2] \) denotes the \(j\)-th RSMB. The feature \( F_{SA} \) initially integrates spatial and angular domain information, providing a foundation for feature refinement and enhancement in Stage II.
\subsubsection{Stage II -- EPMB and EPTB for deep refinement and enhancement of spatial-angular feature in epipolar plane domain}
EPIs not only capture the spatial structure of edges or textures but also reflect disparity information through slanted lines with different slopes \cite{28}. Therefore, Stage II further refines and enhances ${{F}_{SA}}$ in epipolar plane domain, adaptively incorporating structure and disparity information into spatial-angular feature. To achieve this, we design a dual-branch parallel structure consisting of EPMB and EPTB to generate refined features ${{F}_{SAM}} \in \mathbb{R}^{U \times V \times H \times W \times C}$ and ${{F}_{SAT}} \in \mathbb{R}^{U \times V \times H \times W \times C}$:
\begin{equation}
	\small
	\begin{aligned}
		F_{SAM} &= {{H}_{StageII\text{-}M}}(F_{SA})\\ 
		&= EPMB^3(EPMB^2(EPMB^1(F_{SA}))) + F_{SA}, \\
		F_{SAT} &= {{H}_{StageII\text{-}T}}(F_{SA})\\
		&= EPTB^3(EPTB^2(EPTB^1(F_{SA}))) + F_{SA},
	\end{aligned}
\end{equation}
where $EPMB^m(\cdot), m \in [1, 2, 3]$ denotes the \(m\)-th EPMB, and $EPTB^n(\cdot), n \in [1, 2, 3]$ denotes the \(n\)-th EPTB.
\\\indent
EPMB is shown in Fig. \ref{fig3}(d), similar to the structure of SA-RSMB, we first reshape the feature ${{F}_{SA}}$ from Stage I into horizontal EPI form ${{F}_{EPI\_H}} \in \mathbb{R}^{UH \times V \times W \times C}$ and vertical EPI form ${{F}_{EPI\_V}} \in \mathbb{R}^{VW \times U \times H \times C}$. Then, through RSMBs and convolutions, the disparity information in both horizontal and vertical directions interacts within epipolar plane domain. Notably, only two parameter-shared RSMBs and two convolutions are used in EPMB, which further optimizes computational resources:
\begin{equation}
	\begin{aligned}
		F_{EPI\_V} &= Conv(RSMB(F_{EPI\_H})) + F_{EPI\_H}, \\
		F_{SAM} &= Conv(RSMB(F_{EPI\_V})) + F_{EPI\_V}.
	\end{aligned}
\end{equation}
\begin{table*}[]
	\vspace{-0.3cm}
	\renewcommand\arraystretch{1.2} 
	\centering
	\caption{Quantitative comparison (PSNR/SSIM) of different methods for $\times$ 2 and $\times$ 4 LFSR. The best and second-best results are marked in \textcolor{red}{red} and \textcolor[HTML]{3166FF}{blue}. The Params. and FLOPs are calculated on an input LF image with size 5 $\times$ 5 $\times$ 32 $\times$ 32}
	\label{table1}
	\resizebox{\textwidth}{!}{  
		\begin{tabular}{l|c|ccccccc|c}
			\toprule &  &  & 
			\multicolumn{1}{c|}{} & EPFL & HCInew  & HCIold & INRIA & STFgantry & Average                            \\ \cmidrule(l){5-10} 
			\multirow{-2}{*}{Method} & \multirow{-2}{*}{Scale} & \multirow{-2}{*}{Params.(M)} & \multicolumn{1}{c|}{\multirow{-2}{*}{FLOPs(G)}} & PSNR/SSIM                          & PSNR/SSIM                          & PSNR/SSIM                          & PSNR/SSIM                          & PSNR/SSIM                          & PSNR/SSIM                          \\ \midrule
			Bicubic                  & \(\times\) 2                     & -                            & -                                               & 29.50/0.9350                        & 31.69/0.9335                        & 37.46/0.9776                        & 31.10/0.9563                        & 30.82/0.9473                        & 31.11/0.9542                        \\
			EDSR\cite{57}                     & \(\times\) 2                      & 38.62                         & \multicolumn{1}{c|}{989}                        & 33.09/0.9631                        & 34.83/0.9594                        & 41.01/0.9875                        & 34.97/0.9765                        & 36.29/0.9819                        & 36.04/0.9737                        \\
			RCAN\cite{32}                     & \(\times\) 2                      & 15.31                         & \multicolumn{1}{c|}{389.75}                     & 33.16/0.9635                        & 34.98/00.9602                        & 41.05/0.9875                        & 35.01/0.9769                        & 36.33/0.9825                        & 36.11/0.9742                        \\
			resLF\cite{11}                    & \(\times\) 2                      & 7.98                         & \multicolumn{1}{c|}{37.06}                      & 32.75/0.9672                        & 36.07/0.9715                        & 42.61/0.9922                        & 34.57/0.9784                        & 36.89/0.9873                        & 36.58/0.9793                        \\
			LFSSR\cite{12}                    & \(\times\) 2                     & 0.88                         & \multicolumn{1}{c|}{25.70}                      & 33.69/0.9748                        & 36.86/0.9753                        & 43.75/0.9939                        & 35.27/0.9834                        & 38.07/0.9902                        & 37.73/0.9835                        \\
			LF-InterNet\cite{18}              & \(\times\) 2                      & 5.04                         & \multicolumn{1}{c|}{47.46}                      & 34.14/0.9761                        & 37.28/0.9769                        & 44.45/0.9945                        & 35.80/0.9846                        & 38.72/0.9916                        & 38.08/0.9847                        \\
			LF-ATO\cite{13}                   & \(\times\) 2                      & 1.22                         & \multicolumn{1}{c|}{597.66}                     & 34.27/0.9757                        & 37.24/0.9767                        & 44.20/0.9942                        & 36.15/0.9842                        & 39.64/0.9929                        & 38.15/0.9843                        \\
			LF-IINet\cite{17}                 & \(\times\) 2                      & 4.84                         & \multicolumn{1}{c|}{56.16}                      & 34.68/0.9773                        & 37.74/0.9790                        & 44.84/0.9948                        & 36.57/0.9853                        & 39.86/0.9936                        & 38.74/0.9857                        \\
			DistgSSR\cite{19}                 & \(\times\) 2                      & 3.53                         & \multicolumn{1}{c|}{64.11}                      & 34.81/0.9787                        & 37.96/0.9796                        & 44.94/0.9949                        & 36.59/0.9859                        & 40.40/0.9942                        & 38.94/0.9866                        \\
			HLFSR\cite{23}                    & \(\times\) 2                      & 13.72                        & \multicolumn{1}{c|}{167.40}                     & 35.31/0.9800                        & 38.32/0.9807                        & 44.98/0.9950                        & 37.06/0.9867                        & 40.85/0.9947                        & 39.30/0.9874                        \\
			LFT\cite{24}                      & \(\times\) 2                      & 1.11                         & \multicolumn{1}{c|}{56.16}                      & 34.80/0.9781                        & 37.84/0.9791                        & 44.52/0.9945                        & 36.59/00.9855                        & 40.51/0.9941                        & 38.85/0.9863                        \\
			DPT\cite{25}                      & \(\times\) 2                      & 3.73                         & \multicolumn{1}{c|}{65.34}                      & 34.48/0.9758                        & 37.35/0.9771                        & 44.31/0.9943                        & 36.40/0.9843                        & 39.52/0.9926                        & 38.40/0.9848                        \\
			EPIT\cite{28}                     & \(\times\) 2                      & 1.42                         & \multicolumn{1}{c|}{69.71}                      & 34.83/0.9775                        & 38.23/0.9810                        & {\color[HTML]{3166FF} 45.08}/0.9949                        & 36.67/0.9853                        & {\color[HTML]{3166FF} 42.17/0.9957}                        & 39.40/0.9877                        \\
			LF-DET\cite{27}                   & \(\times\) 2                      & 1.59                         & \multicolumn{1}{c|}{48.50}                      & 35.26/0.9797                        & 38.31/0.9807                        & 44.99/0.9950                        & 36.95/0.9864                        & 41.76/0.9955                        & 39.45/0.9875                        \\
			MLFSR\cite{29}                    & \(\times\) 2                      & 1.36                         & \multicolumn{1}{c|}{-}                          & 35.22/0.9801                        & 38.14/0.9803                        & 44.90/0.9950                        & 36.92/0.9865                        & 40.98/0.9949                        & 39.23/0.9874                        \\ \midrule
			LFMT                     & \(\times\) 2                      & 2.06                         & \multicolumn{1}{c|}{63.18}                      & {\color[HTML]{3166FF} 35.52/0.9812} & {\color[HTML]{3166FF} 38.46/0.9815} & {45.03/\color[HTML]{3166FF} 0.9950} & {\color[HTML]{3166FF} 37.21/0.9871} & { 41.92/0.9956} & {\color[HTML]{3166FF} 39.63/0.9881} \\
			LFMT*                    & \(\times\) 2                      & 2.06                         & \multicolumn{1}{c|}{63.18}                      & {\color[HTML]{FE0000} 35.65/0.9819} & {\color[HTML]{FE0000} 38.67/0.9821} & {\color[HTML]{FE0000} 45.31/0.9952} & {\color[HTML]{FE0000} 37.30/0.9875} & {\color[HTML]{FE0000} 42.30/0.9959} & {\color[HTML]{FE0000} 39.85/0.9885} \\ \midrule
			Bicubic                  & \(\times\) 4                       & -                            & \multicolumn{1}{c|}{-}                          & 25.26/0.8324                        & 27.72/0.8517                        & 32.58/0.9344                        & 26.95/0.8867                        & 26.09/0.8452                        & 27.72/0.8701                        \\
			EDSR\cite{57}                     & \(\times\) 4                      & 38.89                         & \multicolumn{1}{c|}{1017}                       & 27.84/0.8854                        & 29.60/0.8869                        & 35.18/0.9536                        & 29.66/0.9257                        & 28.70/0.9072                        & 30.20/0.9118                        \\
			RCAN\cite{32}                     & \(\times\) 4                      & 15.36                         & \multicolumn{1}{c|}{408.5}                      & 27.88/0.8863                        & 29.63/0.8886                        & 35.20/0.9548                        & 29.76/0.9276                        & 28.90/0.9131                        & 30.27/0.9141                        \\
			resLF\cite{11}                    & \(\times\) 4                      & 8.64                         & \multicolumn{1}{c|}{39.70}                      & 28.27/0.9035                        & 30.73/0.9107                        & 36.71/0.9682                        & 30.34/0.9412                        & 30.19/0.9372                        & 31.25/0.9322                        \\
			LFSSR\cite{12}                    & \(\times\) 4                      & 1.77                         & \multicolumn{1}{c|}{128.44}                     & 28.27/0.9118                        & 30.72/0.9145                        & 36.70/0.9696                        & 30.31/0.9467                        & 30.15/0.9426                        & 31.52/0.9370                        \\
			LF-InterNet\cite{18}              & \(\times\) 4                      & 5.48                         & \multicolumn{1}{c|}{50.10}                      & 28.67/0.9162                        & 30.98/0.9161                        & 37.11/0.9716                        & 30.64/0.9491                        & 30.53/0.9409                        & 31.58/0.9388                        \\
			LF-ATO\cite{13}                   & \(\times\) 4                      & 1.36                         & \multicolumn{1}{c|}{686.99}                     & 28.52/0.9115                        & 30.88/0.9135                        & 37.00/0.9699                        & 30.71/0.9484                        & 30.61/0.9430                        & 31.54/0.9373                        \\
			LF-IINet\cite{17}                 & \(\times\) 4                      & 4.88                         & \multicolumn{1}{c|}{57.42}                      & 29.11/0.9188                        & 31.36/0.9208                        & 37.62/0.9734                        & 31.08/0.9515                        & 31.21/0.9502                        & 32.08/0.9429                        \\
			DistgSSR\cite{19}                 & \(\times\) 4                      & 3.58                         & \multicolumn{1}{c|}{65.41}                      & 28.99/0.9195                        & 31.38/0.9217                        & 37.56/0.9732                        & 30.99/0.9519                        & 31.65/0.9535                        & 32.11/0.9440                        \\
			HLFSR\cite{23}                    & \(\times\) 4                      & 13.87                        & \multicolumn{1}{c|}{182.52}                     & 29.20/0.9222                        & 31.57/0.9238                        & 37.78/0.9742                        & 31.24/0.9534                        & 31.64/0.9537                        & 32.29/0.9455                        \\
			LFT\cite{24}                      & \(\times\) 4                      & 1.16                         & \multicolumn{1}{c|}{57.60}                      & 29.25/0.9210                        & 31.46/0.9218                        & 37.63/0.9735                        & 31.20/0.9524                        & 31.86/0.9548                        & 32.28/0.9447                        \\
			DPT\cite{25}                      & \(\times\) 4                      & 3.78                         & \multicolumn{1}{c|}{66.55}                      & 28.93/0.9170                        & 31.19/0.9188                        & 37.39/0.9721                        & 30.96/0.9503                        & 31.14/0.9488                        & 31.93/0.9414                        \\
			EPIT\cite{28}                     & \(\times\) 4                      & 1.47                         & \multicolumn{1}{c|}{74.96}                      & 29.34/0.9197                        & 31.51/0.9231                        & 37.68/0.9737                        & 31.37/0.9526                        & 32.18/0.9571                        & 32.40/0.9452                        \\
			LF-DET\cite{27}                   & \(\times\) 4                      & 1.69                         & \multicolumn{1}{c|}{51.20}                      & 29.47/0.9230                        & 31.56/0.9235                        & {\color[HTML]{3166FF}37.84}/0.9744                        & 31.39/0.9534                        & 32.14/0.9573                        & 32.48/0.9463                        \\
			MLFSR\cite{29}                   & \(\times\) 4                     & 1.41                         & \multicolumn{1}{c|}{-}                          & 29.28/0.9218                        & 31.56/0.9235                        & 37.83/{\color[HTML]{3166FF}0.9745}                        & 31.24/0.9531                        & 32.03/0.9567                        & 32.39/0.9459                        \\ \midrule
			LFMT                     & \(\times\) 4                      & 2.19                         & \multicolumn{1}{c|}{66.72}                      & {\color[HTML]{3166FF} 29.74/0.9239} & {\color[HTML]{3166FF} 31.69/0.9251} & {37.83/0.9744} & {\color[HTML]{3166FF} 31.82/0.9544} & {\color[HTML]{3166FF} 32.23/0.9577} & {\color[HTML]{3166FF} 32.66/0.9471} \\
			LFMT*                    & \(\times\) 4                      & 2.19                         & \multicolumn{1}{c|}{66.72}                      & {\color[HTML]{FE0000} 29.83/0.9255} & {\color[HTML]{FE0000} 31.86/0.9268} & {\color[HTML]{FE0000} 38.01/0.9750} & {\color[HTML]{FE0000} 31.89/0.9553} & {\color[HTML]{FE0000} 32.50/0.9595} & {\color[HTML]{FE0000} 32.82/0.9484} \\ \bottomrule
		\end{tabular}}
	\begin{tablenotes}
		\footnotesize
		\item Notes: * denotes geometry assembling strategy.
	\end{tablenotes}  
	\vspace{-0.5cm}
\end{table*}
\indent
The EPMB is capable of capturing disparity and structural feature in epipolar plane domain. However, compressing these features into limited hidden states may lead to insufficient refinement and enhancement of spatial-angular feature. Therefore, we introduce the Transformer-based EPTB. It has two advantages: (1) compared to feature interaction in spatial or angular domains, the sequence formed by pixel unfolding in epipolar plane domain is relatively short, thus not significantly increasing the computational complexity; (2) experiments show that features extracted by EPTB and EPMB are complementary to each other. The joint use of both can fully leverage their respective advantages, enhancing the richness of feature representations. The structure of EPTB is shown in Fig. \ref{fig3}(e). In this block, we replace the RSMBs in EPMB with Simple Transformer Block (STB):
\begin{equation}
	\begin{aligned}
		F_{EPI\_V} &= Conv(STB(F_{EPI\_H})) + F_{EPI\_H}, \\
		F_{SAM} &= Conv(STB(F_{EPI\_V})) + F_{EPI\_V}.
	\end{aligned}
\end{equation}
\\\indent
Specifically, taking feature $F_{EPI\_H} \in \mathbb{R}^{UH \times V \times W \times C}$ as an example, we first reshape $F_{EPI\_H}$ into an embedding sequence $x \in \mathbb{R}^{UH \times VW \times C}$, which serves as the input to STB. $Q, K, V$ are generated through three linear projection layers ${W}_{Q},{W}_{K},{W}_{V} \in \mathbb{R}^{C \times C}$:
\begin{equation}
	Q = x{W}_{Q}, \quad K = x{W}_{K}, \quad V = x{W}_{V},
\end{equation}
where $Q, K, V$ represent the query, key, and value matrices.
\\\indent
Similar to \cite{60}, Self-Attention (SA) is performed in $P$ independent subspaces to compute Multi-head Self-Attention (MSA):
\begin{equation}
	SA_{p}(Q_{p}, K_{p}, V_{p}) = \text{softmax}\left( \frac{Q_{p} (K_{p})^T}{\sqrt{C/P}} \right) V_{p},
\end{equation}
where $p = 1, 2, \dots, P$. The results from all self-attention heads are concatenated along the channel dimension and projected to produce the weighted sequence $\overline{x} \in \mathbb{R}^{UH \times VW \times C}$:
\begin{equation}
	\overline{x} = Concat(SA_{1}, SA_{2}, \dots, SA_{P}){W}_{P} + x,
\end{equation}
where ${W}_{P} \in \mathbb{R}^{C \times C}$ is the linear projection matrix.
\\\indent
Furthermore, we adopt a feed-forward network layer to introduce local attributes and pixel position awareness, enhancing the spatial information perception capability of STB:
\begin{equation}
	\widehat{x} = MLP(DWConv(LN(\overline{x}))) + \overline{x},
\end{equation}
where \(DWConv(\cdot)\) is the depth-wise convolution, and $\widehat{x} \in \mathbb{R}^{UH \times V \times W \times C}$ is the final output sequence, which adaptively incorporates the horizontal disparity structure. Next, the same processing steps are applied to $F_{EPI\_V} \in \mathbb{R}^{VW \times U \times H \times C}$, incorporating the vertical disparity structure.
\begin{figure*}[htp] 
	\centering
	\vspace{-0.5cm} 
	\includegraphics[width=1\linewidth]{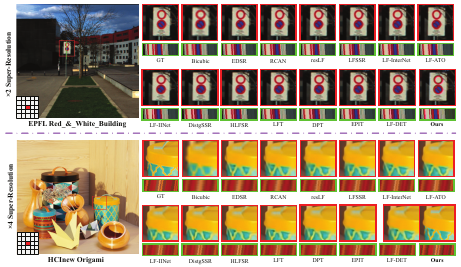}
	\vspace{-0.2cm}
	\caption{Comparison of visual results for the \(\times\) 2 and \(\times\) 4 SR with state-of-the-art methods. As shown in the figure, the enlarged spatial patches of the center SAIs are highlighted with red boxes, and the enlarged EPI slices are highlighted with green boxes.}
	\label{fig5}
	\vspace{-0.5cm}
\end{figure*}

\section{Experiments} \label{Section4}

\subsection{Experiment Settings}
\subsubsection{Datasets}
To ensure fairness and comparability, we use five widely adopted LFSR benchmark datasets: EPFL \cite{52}, HCIold \cite{53}, HCInew \cite{54}, INRIA \cite{55}, and STFgantry \cite{56}, based on settings from previous works \cite{15,19,27}. These datasets contain 144 training scenes and 23 test scenes, covering diverse content and disparity distributions. We extract 5 $\times$ 5 center view images to crop them into high-resolution patches (32 $\times$ 32 and 64 $\times$ 64), and generate corresponding low-resolution patches (16 $\times$ 16) using bicubic interpolation for $\times$ 2 and $\times$ 4 LFSR tasks.
\subsubsection{Implementation Details}
All experiments are conducted on a single NVIDIA RTX A6000 GPU using the PyTorch framework. The LFMT network is trained with $L1$ loss and optimized using Adam with $\beta_{1}=0.9$ and $\beta_{2}=0.999$. The batch size is 2, with an initial learning rate of 2$\times 10^{-4}$, halved every 15 epochs. Training runs for 90 epochs.
\subsubsection{Baseline Methods and Metrics}
We compare our method with several state-of-the-art SR methods, including three single-image methods: Bicubic interpolation, EDSR \cite{57}, and RCAN \cite{32}; seven CNN-based LFSR methods: resLF \cite{11}, LFSSR \cite{12}, LF-InterNet \cite{18}, LF-ATO \cite{13}, LF-IINet \cite{17}, DistgSSR \cite{19}, and HLFSR \cite{23}; four Transformer-based methods: LFT \cite{24}, DPT \cite{25}, EPIT \cite{28}, and LF-DET \cite{27}; and one Mamba-based method: MLFSR \cite{29}. To evaluate image quality, we use two widely adopted metrics: Peak Signal-to-Noise Ratio (PSNR) and Structural Similarity Index (SSIM).
\subsection{Comparison to State-of-the-Arts}
\subsubsection{Quantitative Results}
\begin{figure*}[htb]
	\centering
	\setlength{\abovecaptionskip}{-0.5cm}
	\vspace{-0.1cm}
	\includegraphics[scale=1]{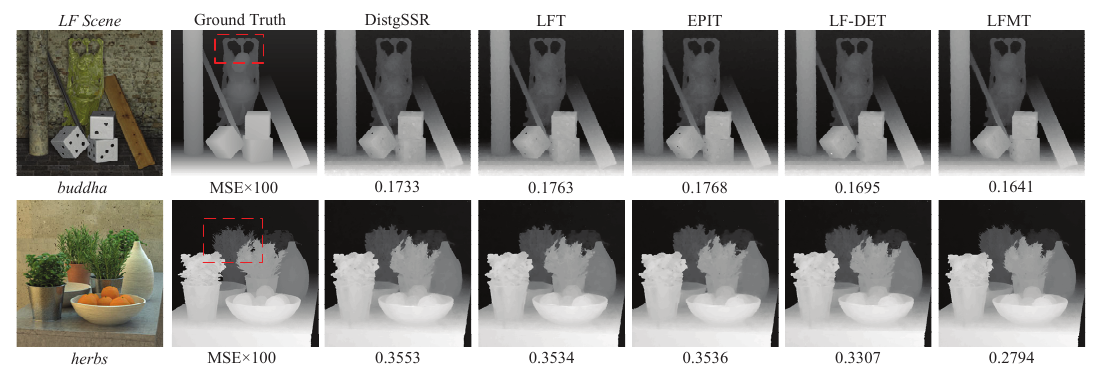}
	\caption{Depth estimation results achieved by SPO \cite{59} using \(\times\) 4 SR LF images generated by different LFSR methods. The mean square error multiplied by 100 (\(MSE\times 100\)) is chosen as the quantitative metric.}
	\label{fig6}
	\vspace{-0.2cm}
\end{figure*}
\begin{figure*}[htp] 
	\centering
	\includegraphics[width=1\linewidth]{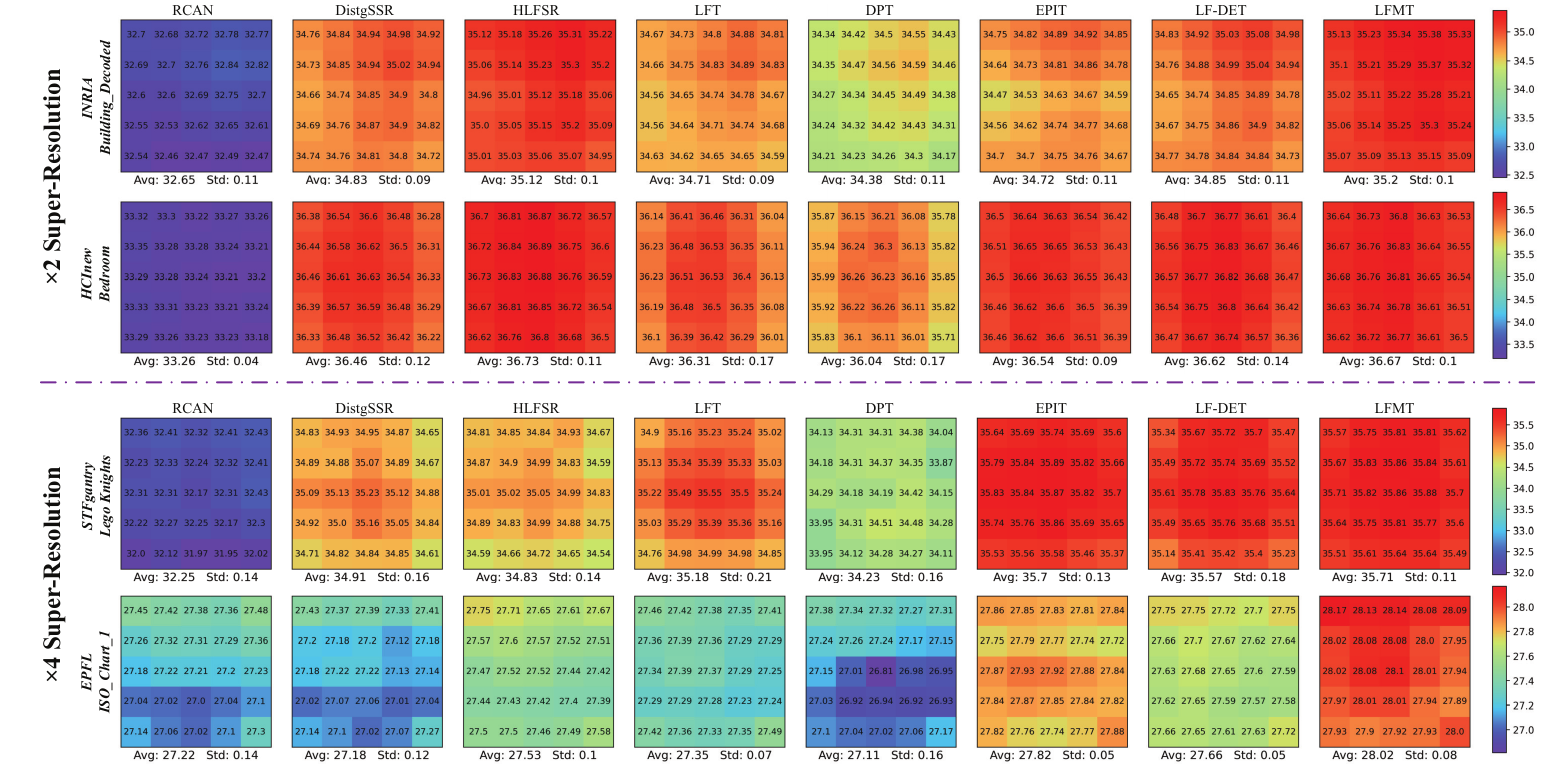}
	\caption{Comparison in terms of PSNR distribution among different views for \(\times\) 2 and \(\times\) 4 SR. Average PSNR and standard deviation for each scene are also listed.}
	\label{fig7}
	\vspace{-0.3cm}
\end{figure*}
Table \ref{table1} presents the quantitative comparison results of LFMT with several advanced methods. On both $\times$ 2 and $\times$ 4 LFSR tasks, our method demonstrates superior performance, significantly outperforming CNN-based and Transformer-based methods, while maintaining a moderate model complexity. Specifically, for $\times$ 2 LFSR, our method achieves the highest average PSNR and SSIM, with a 0.18 dB improvement in average PSNR over the second-best method LF-DET, showcasing excellent reconstruction capability. For the more challenging $\times$ 4 LFSR, our method also shows significant improvements in PSNR and SSIM, with a 0.18 dB and 0.008 increase over LF-DET, respectively. Notably, on real-world datasets such as EPFL and INRIA, our method perform well in modeling complex LF structures and image details, further validating its robustness and broad applicability in real-world scenarios.\\\indent
In-depth performance analysis shows that our method exhibits clear advantages on datasets with large disparities, highlighting its exceptional ability to model long-range correlations in LF images. For example, for the $\times$ 4 LFSR, on the STFgantry dataset, our method improves 0.05 dB in PSNR over the EPIT which excels in large disparity scenes. Meanwhile, the strong performance of our method on mid-to-small disparity datasets, such as the HCInew dataset, highlights its strong compatibility and adaptability in local feature extraction. It is worth noting that on the HCIold dataset, our method slightly lags behind LF-DET, which excels in small disparity and smooth regions due to its multi-scale angular modeling. However, our approach leverages comprehensive spatial, angular, and epipolar interaction to build non-local angular correlations from coarse to fine levels. This holistic modeling improves feature interaction efficiency across subspaces and excels in handling large disparities, complex structures, and artifact reduction. Furthermore, we employed a geometric assembly strategy to further improve overall performance.
\subsubsection{Qualitative Results}
As shown in Fig. \ref{fig5}, for \(\times\) 2 SR our method effectively restores image details, particularly excelling in texture recovery. For instance, in the \(\textit{Red\_ \& White\_Building}\) scene, our method successfully preserves the fine texture of the parking sign and recovers more details in the vertical EPI region. Moreover, our method demonstrates significant advantages in utilizing LF angular information, effectively modeling the spatial-angular correlations, especially in scenes with repetitive textures. For the more challenging \(\times \) 4 SR, our method demonstrates a significant advantage in detail recovery and edge sharpening. For example, in the \(\textit{Origami}\) scene, our method precisely recovers the edges of the crossed lines on the surface of can, outperforming other methods in avoiding common blurring and artifacts, demonstrating excellent detail restoration capability. Due to LFMT, which can integrate multi-level information from spatial, angular, and epipolar plane domains, it shows unique advantages in modeling long-range and fine-grained angular relationships, as demonstrated in the recovery of the horizontal EPI region in \(\textit{Origami}\) scene.
\subsubsection{Angular Consistency}
We conduct analysis from two aspects to evaluate the angular consistency of LFMT. First, we visualize the horizontal or vertical EPI slices from different scenes, as shown in Fig. \ref{fig5}. The EPI slices processed by our method exhibit smoother slanted line textures, outperforming the results of other LFSR methods. For instance, in the \(\textit{Origami}\) scene, the EPI slices generated by LFMT have clearer textures and effectively avoid artifacts, demonstrating higher angular consistency. Second, we perform depth estimation using the SPO algorithm proposed in \cite{59}, selecting the \(\textit{Buddha}\) and \(\textit{Herbs}\) scenes as test objectives. As shown in Fig. \ref{fig6}, the depth maps generated by LFMT have sharper edge textures and lower angular deviation. Our method achieves the best scores in both scenes, further proving its superior angular consistency.
\begin{table}[t]
	\vspace{-0.2cm} 
	\renewcommand\arraystretch{1.2} 
	\centering
	\caption{Comparison of Parameters and FLOPs Between LFMT and Competing Methods on \(\times\) 4 Super-Resolution}
	\label{table2}
	\setlength{\tabcolsep}{3.5mm}{  
		\begin{tabular}{c|c|cc|c}
			\toprule
			Method        & Scale & Params.(M) & FLOPs(G) & PSNR  \\ \midrule
			DistgSSR      & \(\times\) 4    & 3.58       & 65.41    & 32.11 \\
			HLFSR         & \(\times\) 4    & 13.87      & 182.52   & 32.29 \\
			LFT           & \(\times\) 4    & 1.16       & 57.60    & 32.28 \\
			DPT           & \(\times\) 4    & 3.78       & 66.55    & 31.93 \\
			EPIT          & \(\times\) 4    & 1.47       & 74.96    & 32.40 \\
			LF-DET        & \(\times\) 4    & 1.69       & 51.20    & 32.48 \\
			LFMT   (tiny) & \(\times\) 4    & 1.37       & 42.33    & 32.48 \\
			\textbf{LFMT}          & \textbf{\(\times\) 4}    & \textbf{2.19}       & \textbf{66.72}    & \textbf{32.66} \\ \bottomrule
		\end{tabular}
	} 
\end{table}
\begin{figure}[t]
	\centering
	\includegraphics[width=0.48\textwidth,height=0.48\textheight,keepaspectratio]{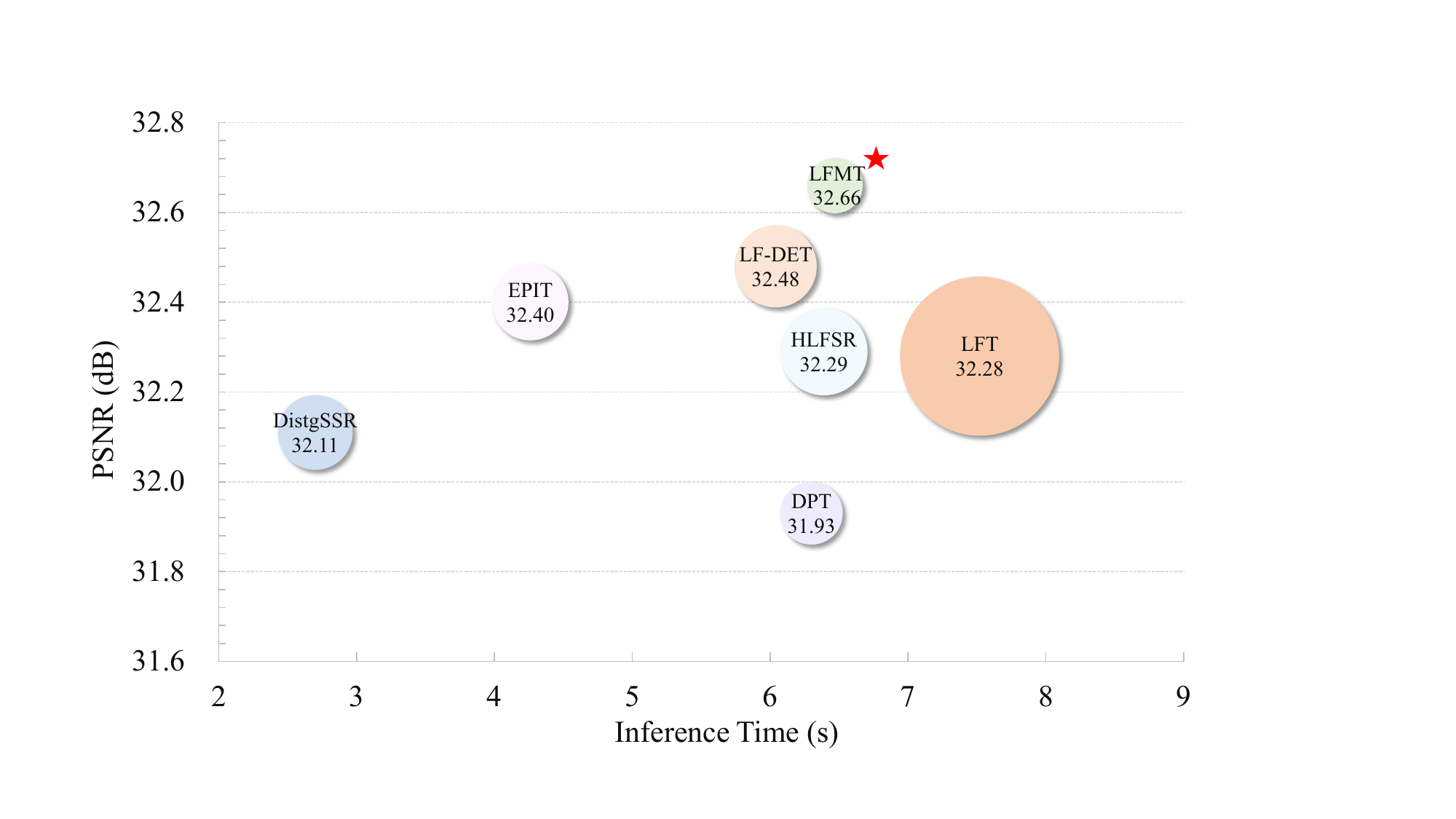}
	\caption{Computational efficiency comparison between LFMT and state-of-the-art methods on \(\times\) 4 SR. The area of circles denotes memory consumption. The inference time is calculated by averaging inference time of all scenes across the five test datasets.}
	\label{fig8}
\end{figure}
\subsubsection{Performance w.r.t. Perspectives}
We further evaluate the reconstruction performance of different methods across different views in the same scene, validating the advantage of LFMT in leveraging non-local spatial-angular information to enhance SR quality. In both \(\times\) 2 and \(\times\) 4 SR tasks, we select several representative scenes with a 5 \(\times\) 5 angular resolution. The PSNR values and standard deviations for each view image are calculated and compared with various advanced methods. As shown in Fig. \ref{fig7}, our method significantly reduces the standard deviation while achieving the highest PSNR values in most scenes. This performance improvement is attributed to the dual-stage non-local spatial-angular correlations modeling strategy, which performs feature extraction in all subspaces. This strategy effectively facilitates the interaction of non-local spatial-angular features in different views, enhancing angular consistency and enabling balanced optimization of  LFSR performance for each view.
\begin{table}[t]
	\vspace{-0.18cm} 
	\renewcommand\arraystretch{1.2} 
	\centering
	\caption{Ablation results (PSNR/SSIM) Comparison of Sub-SS and Other Scanning Strategies for \(\times\) 4 Super-Resolution}
	\label{table3}
	\setlength{\tabcolsep}{6mm}{  
		\begin{tabular}{c|c|c}
			\toprule
			Scanning strategy & Params.(M) & Ave.PSNR/SSIM \\ \midrule
			\textbf{Sub-SS}        & \textbf{2.19}      & \textbf{32.66/0.9471}   \\
			SS2D          & 2.26      & 32.52/0.9468   \\
			Cross-SS2D    & 2.26      & 32.55/0.9470   \\
			SS4D          & 2.34      & 32.52/0.9471    \\ \bottomrule
		\end{tabular}
	} 
\end{table}
\begin{figure}[t]
	\centering
	\includegraphics[width=0.45\textwidth, height=0.45\textheight,keepaspectratio]{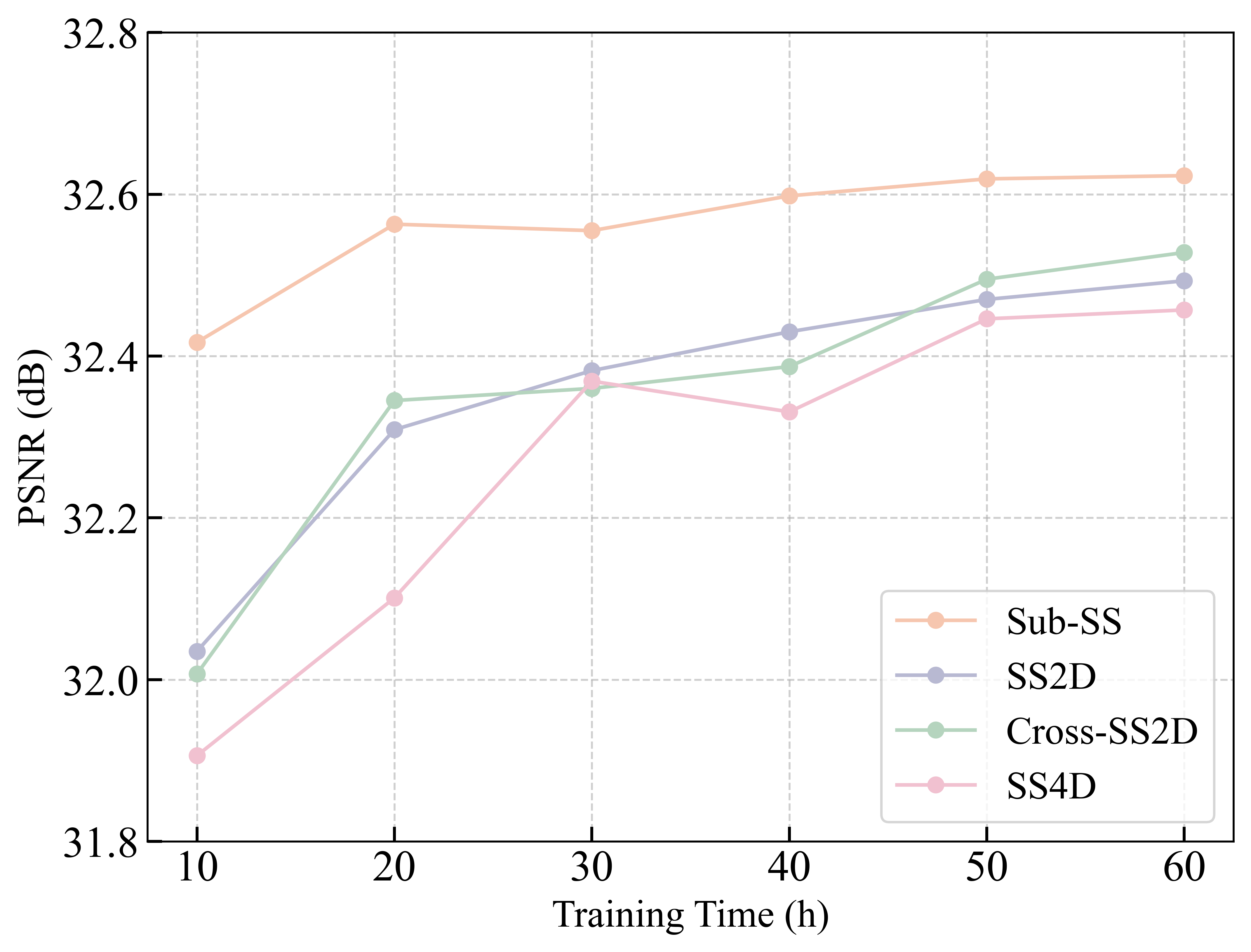}
	\caption{Comparison of PSNR vs. Training Time between Sub-SS and three other scanning strategies with similar parameters on \(\times\) 4 SR.}
	\label{fig9}
\end{figure}
\subsubsection{Computational Efficiency}
To evaluate the computational efficiency of LFMT, we compare its inference time and GPU memory usage with several competitive models, as shown in Fig. \ref{fig8}. In general, compared to CNN-based methods, the inference time of LFMT is slightly increased, while it shows a more competitive inference speed than Transformer-based methods. Notably, LFMT requires less memory consumption while achieving a significant improvement in PSNR. On the other hand, in Table \ref{table2}, we compare the parameters and FLOPs of LFMT with other competing methods. Specifically, we reduced the network depth and trained a tiny LFMT model. Benefiting from the Sub-SS strategy, which reduces redundant computations in subspaces, the tiny LFMT model efficiently simplifies the feature modeling process of LF images. As a result, it achieves the highest PSNR performance with the lowest parameters and FLOPs. Overall, our method strikes a better balance between computational complexity and reconstruction performance, enabling LFMT to efficiently perform LFSR task in resource-constrained hardware environments.
\subsubsection{Limitation}
In the \(\times\) 2 SR, our method does not achieve the highest PSNR on the HCIold \cite{53} and STFgantry \cite{56} datasets. In particular, on STFgantry \cite{56}, it is about 0.25 dB lower than EPIT \cite{28}, likely due to large-disparity structures and locally complex textures, which create a trade-off between local detail recovery and fine-grained angular correlation modeling. Moreover, although LFMT is more computationally efficient than pure Transformer methods, its deployment on resource-constrained devices remains challenging, indicating the need for further model compression and inference optimization.
\begin{figure}[t]
	\centering
	\includegraphics[width=0.5\textwidth, height=0.6\textheight,keepaspectratio]{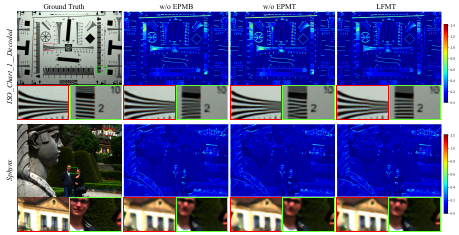}
	\vspace{-0.2cm}
	\caption{Visual results of the reconstructed central view images achieved by different enhancement branches. We also present the error map between the results and ground truth.}
	\label{fig10}
\end{figure}
\begin{table}[t]
	\renewcommand\arraystretch{1.2}
	\centering
	\caption{Ablation results (PSNR/SSIM) on the Effectiveness of SA-RSMB, EPMB, EPTB components for \(\times\) 2 and \(\times\) 4 Super-Resolution}
	\label{table4}
	\setlength{\tabcolsep}{1.2mm}{
		\begin{tabular}{c|c|cc|c} 
			\toprule
			The framework    & Scale & Params.(M) & FLOPs(G) & Ave.PSNR/SSIM \\ \midrule
			LFMT w/o SA-RSMB & \(\times\) 2    & 2.10      & 84.66    & 39.52/0.9878     \\
			LFMT w/o EPMB    & \(\times\) 2    & 2.41      & 64.09    & 39.56/0.9880    \\
			LFMT w/o EPTB    & \(\times\) 2    & 2.37      & 71.16    & 39.12/0.9875    \\
			\textbf{LFMT (ours)}      & \textbf{\(\times\) 2}    & \textbf{2.04}      & \textbf{62.76}    & \textbf{39.63/0.9881}   \\ \bottomrule
			LFMT w/o SA-RSMB & \(\times\) 4    & 2.32      & 86.72    & 32.56/0.9467   \\ 
			LFMT w/o EPMB    & \(\times\) 4    & 2.56      & 68.04    & 32.60/0.9470   \\
			LFMT w/o EPTB    & \(\times\) 4    & 2.15      & 73.79    & 32.46/0.9467   \\
			\textbf{LFMT (ours)}      & \textbf{\(\times\) 4}    & \textbf{2.19}      & \textbf{66.72}    & \textbf{32.66/0.9471}   \\ \bottomrule
	\end{tabular}}
\end{table}

\subsection{Ablation experiments}
In this section, we conduct a series of ablation experiments to investigate the contributions of different components in LFMT network. The experiments include validating the effectiveness of the Sub-SS strategy, SA-RSMB, EPMB, and EPTB components, as well as analyzing the performance of the LFMT network architecture, its hybrid design compared with pure Mamba and pure Transformer models, and its variants. Additionally, we examine the impact of the number of components on model performance and verify the role of hierarchical feature fusion in enhancing LFSR performance. For clarity, the bold entries in the table denote the baseline configuration employed by LFMT.

\subsubsection{The SS-Sub Strategy}
The Sub-SS strategy aims to address the coupling issue present in multi-directional scanning in LF subspaces, thereby improving feature extraction efficiency. To validate the effectiveness of Sub-SS in enhancing computational efficiency while maintaining high performance, we compare it with bidirectional scanning (SS2D \cite{39}), cross-bidirectional scanning (Cross-SS2D) and four-way scanning (SS4D \cite{40}). As shown in Table \ref{table3}, for the \(\times\) 4 SR task, the PSNR values significantly decrease with multi-directional scanning under similar parameter settings. This is because multi-directional scanning leads to redundant modeling of features, causing noise and redundant features to be overly amplified in subspaces, which weakens the recovery of high-frequency details. Further comparisons reveal the advantages of the Sub-SS strategy in training time and model performance. Fig. \ref{fig9} shows the change in the PSNR indicator over time compared with the other three scanning strategies. We observe that under comparable parameter settings, the Sub-SS strategy rapidly converges to a high PSNR value, significantly reducing training time.
\begin{figure}[t]
	\centering
	\includegraphics[width=0.45\textwidth, height=0.5\textheight,keepaspectratio]{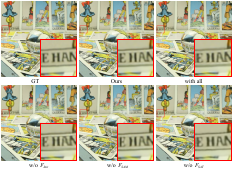}
	\caption{Visualization of the impact of missing features at different levels on the results of \(\times\) 4 SR.}
	\label{fig11}
\end{figure}
\begin{table}[t]
	\renewcommand\arraystretch{1.2}
	\centering
	\caption{Ablation results (PSNR/SSIM) on the Dual-stage Non-local Spatial-angular Correlations Modeling Strategy and Structural Variants for \(\times\) 4 Super-Resolution}
	\label{table5}
	\setlength{\tabcolsep}{0.5mm}{
		\begin{tabular}{c|c|c|c}
			\toprule
			Stage I              & Stage II               & Params. & Ave.PSNR/SSIM \\ \midrule
			\textbf{SA-RSMB}     & \textbf{EPTB+EPMB/Parallel}  
			&   \textbf{2.19}    &\textbf{32.66/0.9471}      \\
			EPTB+EPMB /Parallel & SA-RSMB              &   2.19     &32.44/0.9462               \\ \midrule
			SA-RSMB             & EPMB+EPTB/Sequential &   2.19     &32.59/0.9470       \\
			SA-RSMB             & EPTB+EPMB/Sequential &   2.19     &32.63/0.9471        \\
			\bottomrule
	\end{tabular}}
\end{table}
\begin{table}[t]
	\renewcommand\arraystretch{1.2}
	\centering
	\caption{Ablation results (PSNR/SSIM) of Pure Mamba, Pure Transformer, and LFMT for \(\times\) 4 Super-Resolution}
	\label{table6}
	\setlength{\tabcolsep}{3mm}{
		\begin{tabular}{c|cc|c}
			\toprule
			\multicolumn{1}{l|}{The   framework} & \multicolumn{1}{l}{Params.(M)} & \multicolumn{1}{l|}{FLOPs(G)} & \multicolumn{1}{l}{Ave.PSNR/SSIM} \\ \midrule
			Pure Mamba              & 2.16            & 74.89           & 32.51/0.9468   \\
			Pure Transformer        & 2.23            & 78.62           & 32.54/0.9469   \\
			\textbf{LFMT}           & \textbf{2.19}   &\textbf{66.72}   & \textbf{32.66/0.9471}                  \\
			\bottomrule
	\end{tabular}}
\end{table}
\subsubsection{Effectiveness of SA-RSMB, EPMB, and EPTB Components}
 We evaluate the contributions of the SA-RSMB, EPMB, and EPTB components in the LFMT network by removing each component and retraining the network. To ensure a fair comparison, we adjust the network depth to maintain similar parameters. As shown in Table \ref{table4}, removing any of these components leads to a significant decrease in PSNR values for both \(\times\) 2 and \(\times\) 4 LFSR tasks. This is because SA-RSMB effectively selects and interacts with important features in both spatial and angular domains, while the EPMB and EPTB process and enhance deep spatial-angular features in epipolar plane domain, strengthening spatial-angular correlations. Fig. \ref{fig10} presents visual comparisons  showing the impact of missing EPMB or EPTB components, further confirming the synergistic effect of EPMB and EPTB. These components are crucial for feature extraction and interaction in all subspaces of LF images, significantly contributing to the improvement of LFSR performance.
\subsubsection{Dual-stage Non-local Spatial-angular Correlations Modeling Strategy and Structural Variants}
 We conduct an in-depth ablation study on the structure and variants of the dual-stage non-local spatial-angular correlation modeling in LFMT. As shown in Table \ref{table5}, we first examine the impact of swapping the order of feature extraction in spatial-angular domain and epipolar plane domain. In the baseline LFMT, Stage I employs SA-RSMB for coarse spatial-angular feature extraction, followed by Stage II with EPTB and EPMB for fine epipolar feature enhancement, achieving a PSNR of 32.66 dB. In contrast, the “reversed stage order” experiment first extracts epipolar features with EPTB and EPMB, and then applies SA-RSMB in Stage II for spatial-angular feature extraction, which results in a lower PSNR of 32.44 dB (a drop of 0.22 dB). This comparison demonstrates that performing coarse spatial-angular extraction first, followed by epipolar refinement, is both reasonable and essential. Such a design sequence is more favorable for effectively modeling non-local spatial-angular correlations in LF images. This is because, at deeper network layers, the intertwined spatial-angular information in epipolar plane domain more accurately represents disparity information and inherent structures, thereby facilitating high-quality recovery and modeling of non-local spatial-angular features. Additionally, we compare the performance of sequential and parallel configurations of the EPMB and EPTB components. The results indicate that the parallel configuration better leverages the potential of these components in the spatial-angular feature refinement stage. This configuration promotes the complementary advantages of EPMB and EPTB, effectively enhancing the LFSR performance.
 \begin{table}[t]
 	\renewcommand\arraystretch{1.2}
 	\centering
 	\caption{Ablation results (PSNR/SSIM) on the Number of SA-RSMB, EPMB, EPTB Components for \(\times\) 4 Super-Resolution}
 	\label{table7}
 	\setlength{\tabcolsep}{1.5mm}{
 		\begin{tabular}{@{}c|c|cc|c@{}}
 			\toprule
 			\begin{tabular}[c]{@{}c@{}}Number   of  \\ SA-RSMB\end{tabular} & \begin{tabular}[c]{@{}c@{}}Number of  \\ EPIM/EPIT\end{tabular} & {Params.(M)} & FLOPs(G)       & Ave.PSNR/SSIM        \\ \midrule
 			\textbf{3}                                                      & \textbf{3}                                                      & \textbf{2.19}                  & \textbf{66.72} & \textbf{32.66/0.9471} \\
 			1                                                               & 3                                                               & 1.51                           & 50.11          & 32.46/0.9454          \\
 			2                                                               & 3                                                               & 1.85                           & 58.41          & 32.61/0.9467          \\
 			4                                                               & 3                                                               & 2.53                           & 75.02          & 32.64/0.9470          \\
 			3                                                               & 1                                                               & 1.61                           & 44.26          & 32.35/0.9454          \\
 			3                                                               & 2                                                               & 1.90                           & 55.49          & 32.53/0.9465          \\
 			3                                                               & 4                                                               & 2.48                           & 77.95          & 32.63/0.9470          \\ \bottomrule
 	\end{tabular} }
 \end{table}
 
 \begin{table}[t]
 	\renewcommand\arraystretch{1.2}
 	\centering
 	\caption{Ablation results (PSNR/SSIM) on THE Hierarchical Feature Fusion for \(\times\) 4 Super-Resolution}
 	\label{table8}
 	\setlength{\tabcolsep}{3mm}{
 		\begin{tabular}{c|cc|c}
 			\toprule
 			\multicolumn{1}{l|}{The   framework} & \multicolumn{1}{l}{Params.(M)} & \multicolumn{1}{l|}{FLOPs(G)} & \multicolumn{1}{l}{Ave.PSNR/SSIM} \\ \midrule
 			with all                      & 2.25      & 68.39     & 32.62/0.9471   \\
 			w/o  $F_{\text{Init}}$        & 2.19      & 66.72      &32.60/0.9469   \\
 			\textbf{w/o $F_{\text{SA}}$}  & \textbf{2.19}                          &\textbf{66.72}               & \textbf{32.66/0.9471}                  \\
 			w/o $F_{\text{SAT}}$          & 2.19      & 66.72     & 32.37/0.9457    \\
 			w/o $F_{\text{SAM}}$          & 2.19      & 66.72     & 32.58/0.9467     \\
 			\bottomrule
 	\end{tabular}}
 \end{table}

\subsubsection{Comparison of Pure Mamba, Pure Transformer, and LFMT}
To further validate the complementarity between the Mamba and Transformer mechanisms, we conducted a structural-level comparison among three full model variants: a pure Mamba network, a pure Transformer network, and the proposed hybrid LFMT network. As shown in Table \ref{table6}, the hybrid LFMT network achieves the best overall performance, with PSNR improvements of 0.15 dB over the pure Mamba network and 0.12 dB over the pure Transformer network, significantly outperforming the two single-mechanism networks. We attribute this performance gain to the hybrid design of LFMT, which integrates both state space modeling and self-attention modeling paradigms. This integration effectively enlarges the receptive field and allows more comprehensive modeling of non-local spatial-angular correlations across multiple subspaces in the LF data. By combining the efficient long-range dependency modeling capability of Mamba with the fine-grained feature representation of Transformer, LFMT demonstrates superior reconstruction quality and enhanced generalization. These results fully validate the advantage of the hybrid paradigm in LF modeling.
\subsubsection{Number of SA-RSMB, EPMB, EPTB Components}
\label{dual_progressive}
In Table \ref{table7}, we provide a detailed analysis of the impact of different quantities of SA-RSMB and EPMB/EPTB on the overall performance of LFMT. Notably, the counts of EPMB and EPTB components remain identical. The bolded data in the first row represents the configuration adopted by our method, where the number of SA-RSMB, EPMB and EPTB is set to 3. The experimental results show that, when the number of EPMB/EPTB is fixed at 3, as the number of SA-RSMB increases from 1 to 4, the Ave.PSNR/SSIM first increases and then decreases, reaching its peak when the number of SA-RSMB is 3. Similarly, when the number of SA-RSMB is fixed at 3, as the number of EPMB/EPTB increases from 1 to 4, the Ave.PSNR/SSIM also follows a trend of first increasing and then decreasing, stabilizing when the number of EPMB/EPTB is 3. These results clearly indicate that the number of SA-RSMB and EPMB/EPTB components significantly affects the performance of LFMT. Too few components may result in inadequate feature learning, while too many can lead to overfitting. After considering both model complexity and network performance, we determined that the optimal configuration is to set the number of SA-RSMB, EPMB and EPTB to 3. This configuration ensures efficient operation of the LFMT network while achieving the best LFSR performance.
\subsubsection{Hierarchical Feature Fusion}
We conduct ablation experiments on feature fusion at different layers to investigate the impact of hierarchical feature fusion on model performance. Specifically, we respectively remove $F_{\text{Init}}$, $F_{\text{SA}}$, $F_{\text{SAT}}$, $F_{\text{SAM}}$, and conduct performance evaluations for the ×4 LFSR task. As shown in Table \ref{table8}, we are surprised to find that the fusion of only $F_{\text{Init}}$, $F_{\text{SAT}}$, and $F_{\text{SAM}}$ can maximally unleash the potential of LFMT, achieving the highest performance. In particular, removing $F_{\text{Init}}$ only slightly affects the performance, while adding $F_{\text{SA}}$ leads to a decline in performance. Furthermore, the removal of $F_{\text{SAT}}$ or $F_{\text{SAM}}$ disrupts the interaction of deep features within the epipolar plane domain, thereby weakening the model's capacity to leverage disparity information and accurately model the intrinsic LF structure. This has a significant detrimental impact on overall performance. Fig. \ref{fig11} visualizes the effects of missing features at various levels on the results. These experimental results demonstrate that hierarchical feature fusion plays a crucial role in the LFSR task. And it is essential to ensure that the LFSR network effectively captures and integrates feature information across multiple levels, including spatial, angular, and epipolar plane domains.
\section{Conclusion} \label{Section5}
SSMs have shown promise in visual tasks, but faces challenges in LFSR due to the dense and complex 4D correlations far exceeding those in 2D images. This paper introduces LFMT, a LFSR network that leverages non-local spatial-angular information in complete LF subspaces to showcase the potential of SSMs in LFSR tasks. By incorporating the Sub-SS strategy and optimizing the Mamba structure, LFMT enhances long-range dependency modeling and computational efficiency. Additionally, we propose a dual-stage modeling strategy for deep non-local spatial-angular correlations modeling and combine the feature extraction strengths of SSM and Transformer models to comprehensively explore disparity information and LF intrinsic structure information. Comparative experiments demonstrate that LFMT achieves superior performance over CNN-based and Transformer-based methods, highlighting its ability to utilize complete subspace information effectively. LFMT's interactive modeling of spatial-angular-epipolar subspaces offers a transferable framework for various LF processing tasks. In future work, we will explore extending LFMT to other LF reconstruction and enhancement applications, such as   angular super-resolution.

\begin{footnotesize}
	\bibliographystyle{IEEEtran}
	\bibliography{ref-LFMT}
\end{footnotesize}
\newpage
\vfill
\begin{IEEEbiography}[{\includegraphics[width=1in,height=1.25in,clip,keepaspectratio]{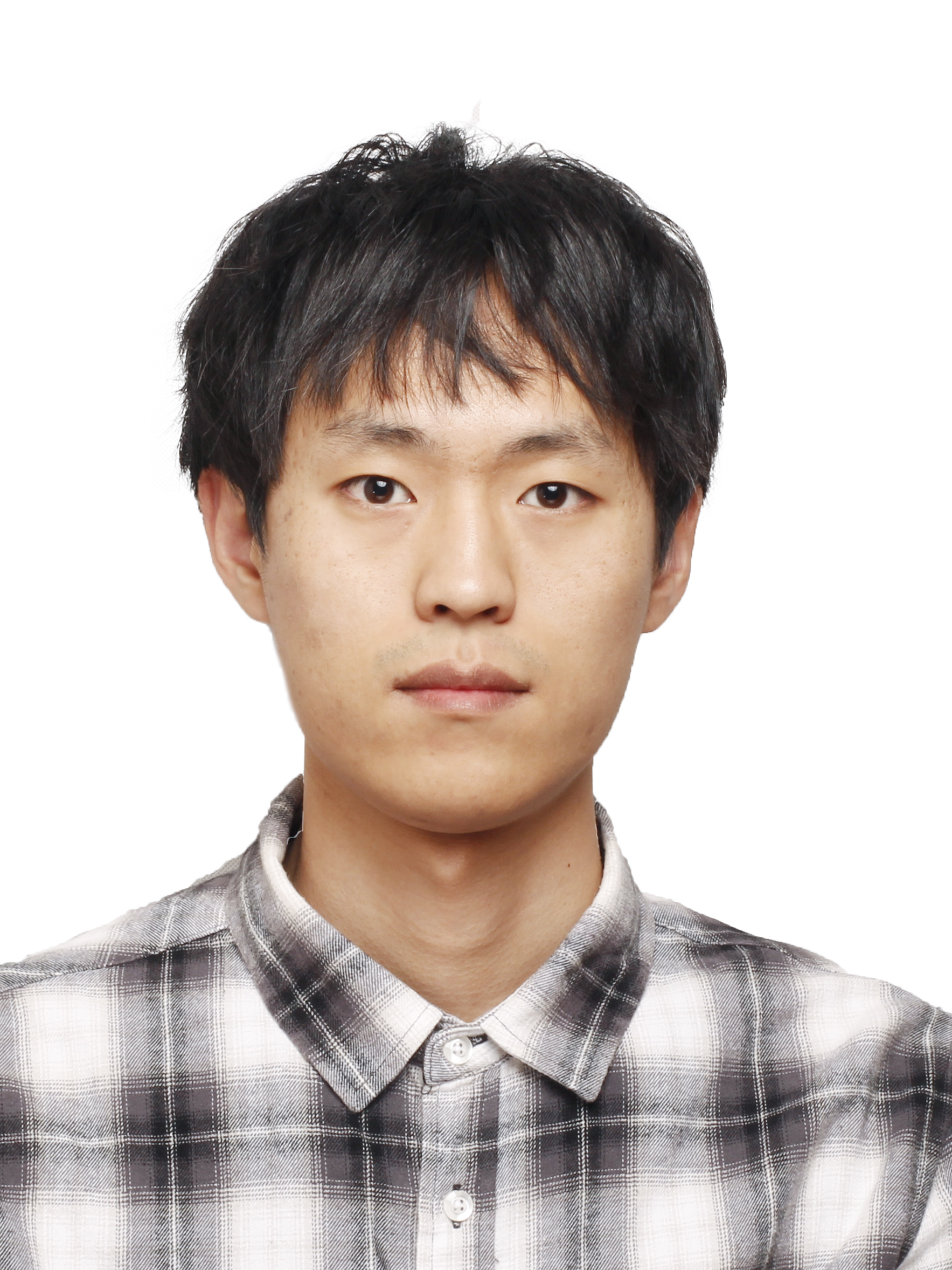}}]{Haosong Liu} received the B.E. degree from the School of Information Science and Engineering, Huaqiao University, Xiamen, China, in 2023. He is currently pursuing the M.S. degree in Information and Communication Engineering with the School of Information Science and Engineering, Huaqiao University, Xiamen, China. His research interests include light field image restoration and deep learning.
\end{IEEEbiography}
\vspace{-1cm}
\begin{IEEEbiography}[{\includegraphics[width=1in,height=1.25in,clip,keepaspectratio]{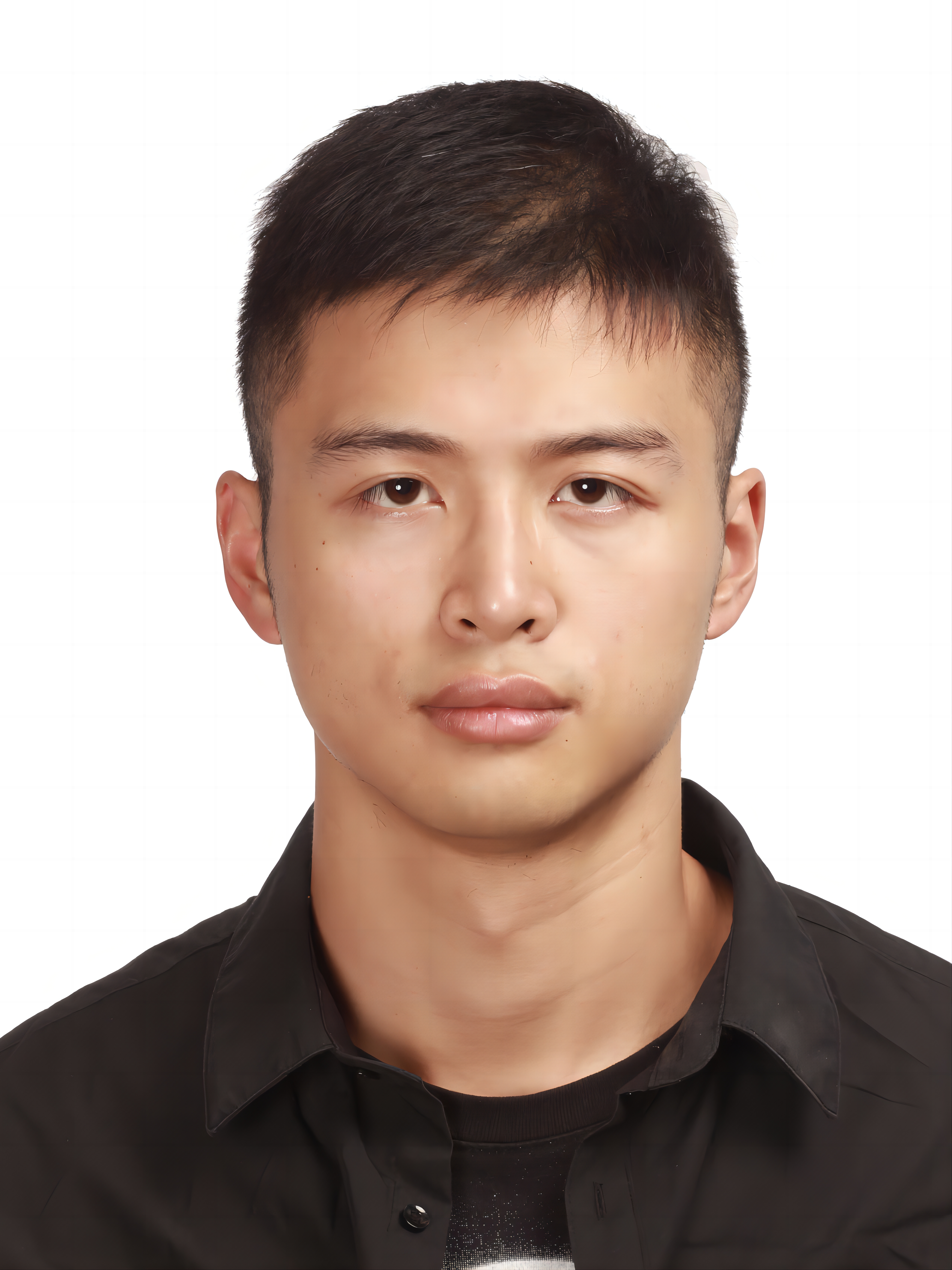}}]{Xiancheng Zhu} received the M.S. degree in computer technology from Huaqiao University, Quanzhou, China, in 2023. He is currently pursuing the Ph.D. degree in intelligent manufacturing engineering at the School of Mechanical Engineering and Automation, Huaqiao University. His research interests include image restoration, industrial image super-resolution, and deep learning.
\end{IEEEbiography}
\vspace{-1cm}
\begin{IEEEbiography}[{\includegraphics[width=1in,height=1.25in,clip,keepaspectratio]{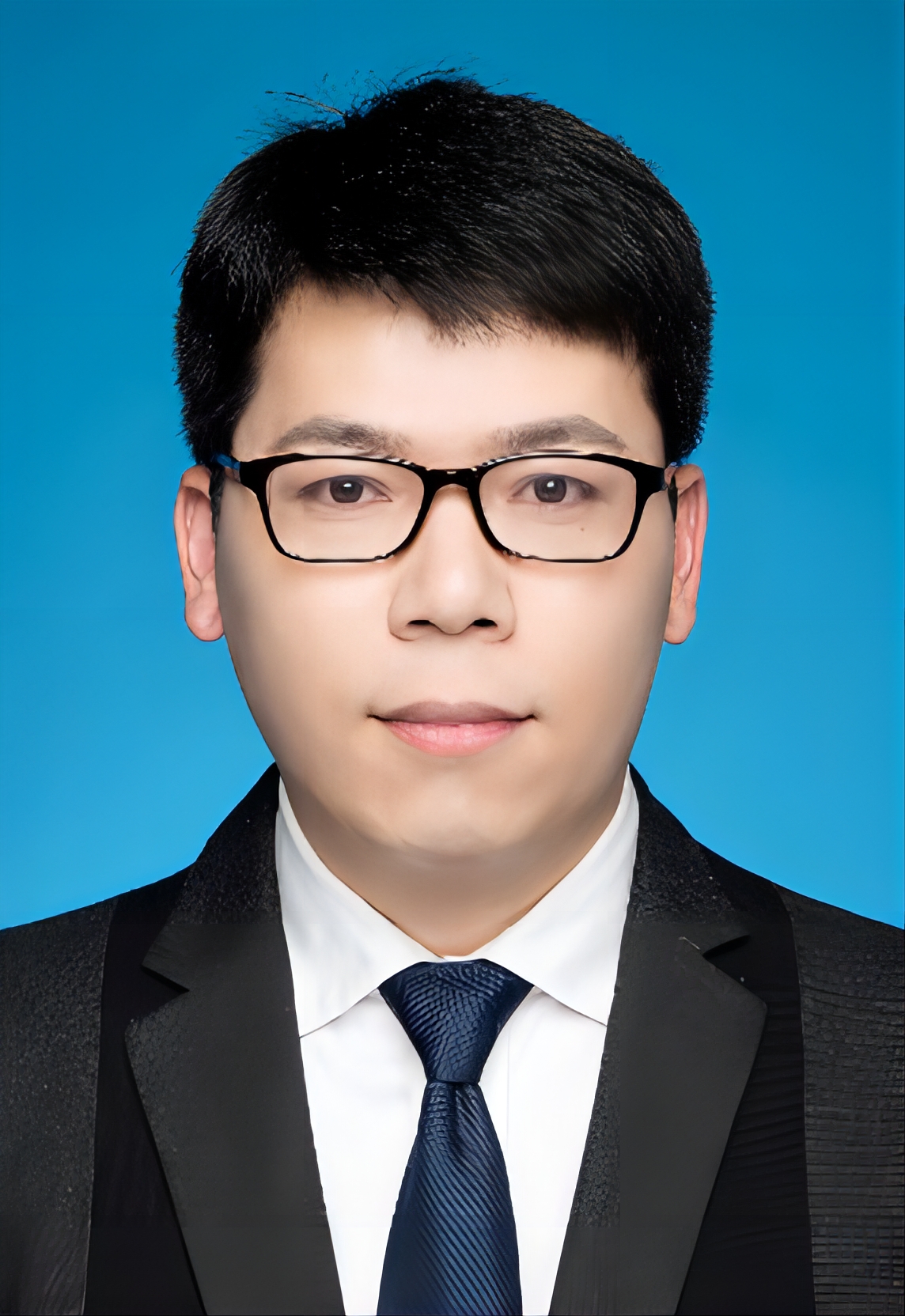}}]{Huanqiang Zeng} (Senior Member, IEEE) received the B.S. and M.S. degrees in electrical engineering from Huaqiao University, China, and the Ph.D. degree in electrical engineering from Nanyang Technological University, Singapore.\\\indent
He is currently a Full Professor of Huaqiao University and Xiamen University of Technology. Before that, he was a Postdoctoral Fellow at The Chinese University of Hong Kong, Hong Kong. He has published more than 190 papers in well-known journals and conferences, including three best poster/paper awards (in the International Forum of Digital TV and Multimedia Communication 2018 and the Chinese Conference on Signal Processing 2017/2019). His research interests include image processing, video coding, machine learning, and computer vision. He has also been actively serving as the General Co-Chair for IEEE International Symposium on Intelligent Signal Processing and Communication Systems 2017 (ISPACS2017), the Co-Organizer for ICME2020 Workshop on 3D Point Cloud Processing, Analysis, Compression, and Communication, the Technical Program Co-Chair for Asia–Pacific Signal and Information Processing Association Annual Summit and Conference 2017 (APSIPA-ASC2017), the Area Chair for IEEE International Conference on Visual Communications and Image Processing (VCIP2015 and VCIP2020), and a technical program committee member for multiple flagship international conferences. He has been actively serving as an Associate Editor for IEEE Transactions on Image Processing, IEEE Transactions on Circuits and Systems for Video Technology, and IET Electronics Letters, and a Senior Area Editor for IEEE Signal Processing Letters.
\end{IEEEbiography}
\vspace{-1cm}
\begin{IEEEbiography}[{\includegraphics[width=1in,height=1.25in,clip,keepaspectratio]{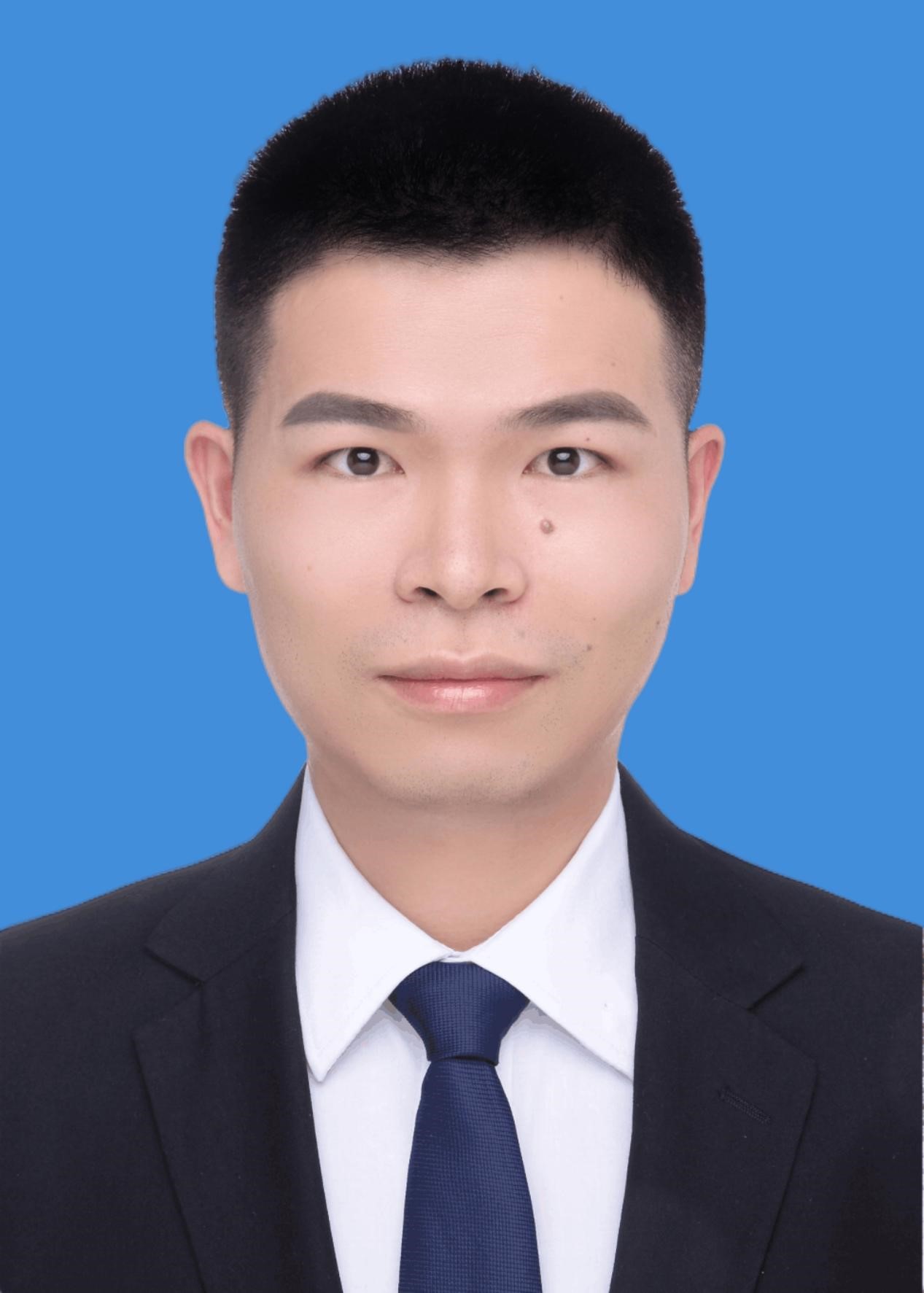}}]{Jianqing Zhu} (Senior Member, IEEE)received the B.S. degree in Communication Engineering and the M.S. degree in Communication and Information System from the School of Information Science and Engineering, Huaqiao University, Xiamen, China, in 2009 and 2012, respectively. He received the Ph.D. degree in Computer Application Technology from the Institute of Automation, Chinese Academy of Sciences, Beijing, China, in 2015. He is now a Professor at the College of Engineering, Huaqiao University, Quanzhou, China. His current research interests include computer vision and pattern recognition, with a focus on image and video analysis, particularly person re-identification, object detection, and video surveillance. He was awarded the Best Biometrics Student Paper award at the International Conference on Biometrics in 2015.
\end{IEEEbiography}

\begin{IEEEbiography}[{\includegraphics[width=1in,height=1.25in,clip,keepaspectratio]{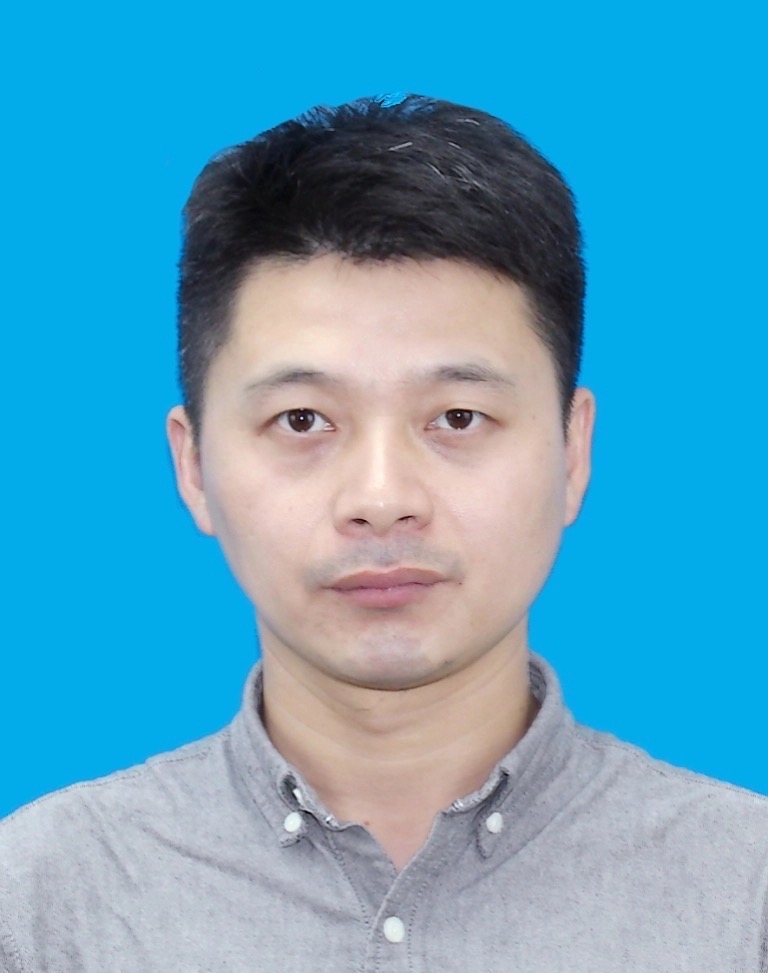}}]{Jiuwen Cao} (Senior Member, IEEE) received the B.Sc. and M.Sc. degrees from the School of Applied Mathematics, University of Electronic Science and Technology of China, Chengdu, China, in 2005 and 2008, respectively, and the Ph.D. degree from the School of Electrical and Electronic Engineering, Nanyang Technological University (NTU), Singapore, in 2013. From 2012 to 2013, he was a Research Fellow at NTU. He is currently a Professor and Vice Dean of the School of Automation at Hangzhou Dianzi University, Hangzhou, China. His main research interests include machine learning, neural networks, and medical signal processing. He has published over 180 papers in top-tier journals and conferences and has been awarded the Best Conference Paper of ICCSIP 2020 and the Best Conference Paper Finalist of ICCSIP 2021. He is an Associate Editor of IEEE Transactions on Circuits and Systems—I: Regular Paper, The Journal of the Franklin Institute, Neurocomputing, Military Medical Research, and Multidimensional Systems and Signal Processing.
\end{IEEEbiography}
\vspace{-7.7cm}
\begin{IEEEbiography}[{\includegraphics[width=1in,height=1.25in,clip,keepaspectratio]{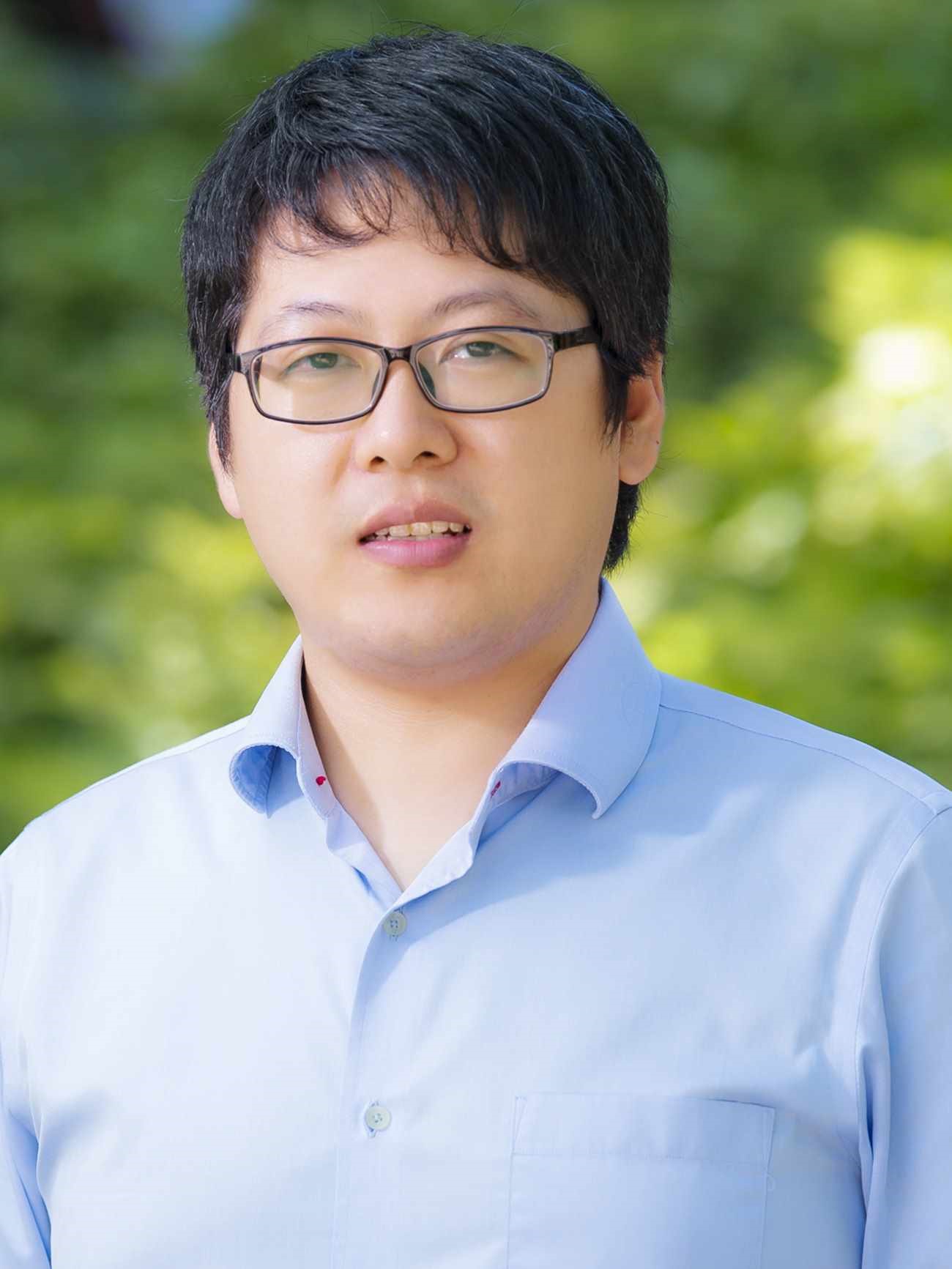}}]{Junhui Hou} is an Associate Professor with the Department of Computer Science, City University of Hong Kong. He holds a B.Eng. degree in information engineering (Talented Students Program) from the South China University of Technology, Guangzhou, China, an M.Eng. degree in signal and information processing from Northwestern Polytechnical University, Xi’an, China, and a Ph.D. degree from the School of Electrical and Electronic Engineering, Nanyang Technological University, Singapore. His research interests are neural spatial computing.\\\indent
Dr. Hou received the Early Career Award (3/381) from the Hong Kong Research Grants Council in 2018 and the NSFC Excellent Young Scientists Fund in 2024. He has served or is serving as an Associate Editor for \textit{IEEE Transactions on Visualization and Computer Graphics}, \textit{IEEE Transactions on Image Processing}, \textit{IEEE Transactions on Multimedia}, and \textit{IEEE Transactions on Circuits and Systems for Video Technology}.
\end{IEEEbiography}
\end{document}